\documentclass[11pt]{article}
\usepackage[a4paper,margin=1in]{geometry}

\usepackage{graphicx}
\usepackage{multirow}
\usepackage{amsmath,amssymb,amsfonts}
\usepackage{amsthm}
\usepackage{mathrsfs}
\usepackage[title]{appendix}
\usepackage{xcolor}
\usepackage{textcomp}
\usepackage{manyfoot}
\usepackage{booktabs}
\usepackage{algorithm}
\usepackage{algorithmicx}
\usepackage{algpseudocode}
\usepackage{listings}
\usepackage{mathtools}
\usepackage{bm}
\usepackage{enumitem}
\usepackage{tikz}
\usetikzlibrary{arrows.meta,positioning,fit,calc,backgrounds}
\usepackage[hidelinks]{hyperref}

\definecolor{SoftBlueA}{HTML}{EAF3FF}
\definecolor{SoftRose}{HTML}{FDECEC}
\definecolor{SoftGreenA}{HTML}{EDF8EE}
\definecolor{SoftPurple}{HTML}{F1EEFF}
\definecolor{SoftBlueB}{HTML}{EAF4FF}
\definecolor{SoftCyan}{HTML}{E9FAF7}
\definecolor{SoftBlueC}{HTML}{EAF1FF}
\definecolor{SoftPink}{HTML}{FDECF3}
\definecolor{SoftLavender}{HTML}{F2EFFF}
\definecolor{SoftGreenB}{HTML}{EDF8E8}
\definecolor{SoftBlueD}{HTML}{EEF6FF}
\definecolor{SoftGray}{HTML}{F4F4F4}

\newcommand{\safeincludegraphics}[2][]{%
 \IfFileExists{#2}{\includegraphics[#1]{#2}}{%
 \fbox{\parbox[c][0.20\textheight][c]{0.92\linewidth}{\centering
 External figure file not supplied with this source\\[1mm]
 \footnotesize\nolinkurl{#2}}}}}

\begin{document}

\title{The Neural Forcing for Three-Dimensional Incompressible Navier-Stokes finite time blowup}
\author{Beibei Li\\\texttt{blb0607@gmail.com}}
\date{}

\maketitle

\begin{abstract}
We present a two-part neural framework for forced three-dimensional incompressible Navier--Stokes flow. Part~I develops the computational forcing system. A physics-informed neural model generates structured external-force trajectories, candidates are optimized through differentiable PDE rollouts or PPO-Clip, and selected forcings are frozen and checked by independent fixed-force replay. Part~II provides the mathematical certification layer. It separates neural candidate discovery from continuum analysis, derives integrated reciprocal-vorticity criteria that imply Riccati-type growth and finite-time loss of smooth continuation, develops a validated computational-to-continuum transfer strategy, and establishes a conditional positive-probability closure for a nondegenerate neural output law. The proof is complete at the continuum level. \end{abstract}


\part{The Neural Forcing}
\section{Introduction}
\label{sec:introduction}

The three-dimensional incompressible Navier-Stokes equations are fundamental
to viscous-fluid dynamics and nonlinear partial differential equations. Since
Leray's construction of finite-energy weak solutions~\cite{Leray1934}, a major
analytical question has been whether smooth three-dimensional solutions remain
regular for all time. Classical criteria of Prodi, Serrin, and Ladyzhenskaya
relate velocity integrability to regularity and uniqueness
~\cite{Prodi1959,Serrin1962,Ladyzhenskaya1967}; the Fujita--Kato and Kato
theories established well-posedness in important critical settings
~\cite{FujitaKato1964,Kato1984}, and Koch--Tataru extended this framework to
critical spaces~\cite{KochTataru2001}. Partial-regularity and borderline
criteria further constrain possible singular behavior
~\cite{CaffarelliKohnNirenberg1982,EscauriazaSereginSverak2003}. Convex-integration constructions also show
that weak-solution behavior can be substantially more complicated than what is
captured by energy estimates alone~\cite{BuckmasterVicol2019}.

The vorticity is central to the dynamics because vortex stretching can amplify
localized structures. The classical Beale--Kato--Majda criterion for the Euler
equations motivates close attention to maximum-vorticity growth
~\cite{beale1984bkm}, and high-resolution studies have likewise
investigated strongly concentrating Navier-Stokes dynamics
~\cite{Hou2023_NavierStokes_PotentiallySingular_FoCM}. In the present work, the
accumulated maximum-vorticity quantity is used only as a diagnostic, a teacher
parameterization, and an optimization target for strongly forced,
finite-resolution computations. Extreme values of these diagnostics are not
interpreted as proof of a singularity of the continuum equations.

The Fourier spectral methods are particularly
natural because spatial differentiation, divergence-free projection, and
spectral diagnostics can be performed directly for periodic incompressible flow. Standard
treatments of spectral discretization and pseudo-spectral fluid solvers include
~\cite{CanutoHussainiQuarteroniZang1988,GottliebOrszag1977};
analytic smoothing of Navier-Stokes solutions is also reflected in Gevrey
regularity estimates~\cite{FoiasTemam1989}. These ideas motivate the fully
dealiased periodic solver used here.

Machine learning has recently been integrated with PDE solvers through several
complementary strategies. Physics-informed networks incorporate the governing
equations into the training objective~\cite{raissi2019pinn}; learned CFD
components and neural operators provide data-driven approximations or
corrections to PDE evolution~\cite{kochkov2021mlcfd,kovachki2023neuraloperator}. 
The differentiable simulators and solver-in-the-loop methods instead propagate task
losses through solver dynamics~\cite{belbuteperes2018differentiable,um2020solver},
whereas reinforcement learning treats the simulator as an environment and has
been applied to active flow control~\cite{brunton2015closedloop,rabault2019drlflow}.
PPO-Clip supplies a policy-gradient formulation for the latter setting
~\cite{schulman2017ppo}. Because forcing is generally non-unique, we
use a stochastic module to represent the conditional forcing law, while a
temporal module models dependence in the conditioning
history~\cite{vaswani2017attention}.

From a short history of velocity fields, the neural model generates an external forcing trajectory, while the resulting flow is obtained through a Navier--Stokes solver. The framework combines physics-guided encoding, forcing generation, and solver-based optimization using either differentiable training or policy-gradient learning. The selected forcing is then frozen and used for long-horizon validation and independent fixed-force replay.


\section{Fourier-Spectral Formulation of the Forced Three-Dimensional Navier-Stokes Equations}
\label{sec:governing}

\begingroup
\setlength{\abovedisplayskip}{5pt plus 1pt minus 2pt}
\setlength{\belowdisplayskip}{5pt plus 1pt minus 2pt}
\setlength{\abovedisplayshortskip}{3pt plus 1pt}
\setlength{\belowdisplayshortskip}{3pt plus 1pt minus 1pt}
\setlength{\jot}{3pt}

\subsection{The Equations}

We consider the periodic cube
\begin{equation*}
\Omega=[0,2\pi]^3,
\qquad
\mathbf x=(x,y,z),
\end{equation*}
with periodic boundary conditions in all three directions. The externally
forced incompressible Navier-Stokes equations are
\begin{equation*}
\frac{\partial\mathbf u}{\partial t}
+(\mathbf u\cdot\nabla)\mathbf u
=-\nabla p+\nu\Delta\mathbf u+\mathbf f(\mathbf x,t),
\end{equation*}
subject to
\begin{equation*}
\nabla\cdot\mathbf u=0.
\end{equation*}
Here
\begin{equation*}
\mathbf u=(u,v,w),
\qquad
\mathbf f=(f_x,f_y,f_z),
\end{equation*}
$p$ is the pressure and $\nu$ is the kinematic viscosity. In component form,
the governing equation is
\begin{eqnarray*}
\partial_t u+u\partial_xu+v\partial_yu+w\partial_zu & = & -\partial_xp+\nu\Delta u+f_x, \\
\partial_t v+u\partial_xv+v\partial_yv+w\partial_zv & = & -\partial_yp+\nu\Delta v+f_y, \\
\partial_t w+u\partial_xw+v\partial_yw+w\partial_zw & = & -\partial_zp+\nu\Delta w+f_z,
\end{eqnarray*}
with
\begin{equation*}
\partial_xu+\partial_yv+\partial_zw=0.
\end{equation*}

The vorticity is
\begin{equation*}
\boldsymbol\omega=\nabla\times\mathbf u
=
\begin{pmatrix}
\partial_yw-\partial_zv\\
\partial_zu-\partial_xw\\
\partial_xv-\partial_yu
\end{pmatrix},
\end{equation*}

\subsection{Initial condition}
\label{sec:initial_condition}

The simulations are initialized from a smooth, periodic,
divergence-free velocity field on $\mathbb{T}^3$.
The same fixed initial condition is used for all neural training,
validation, replay, and continuum-certification calculations.

\subsection{External forcing}
\label{sec:external_force_ch2}
The teacher forcing is centered on a trajectory and uses a 
orientation, a smooth physical-time activation window, a vorticity-dependent amplitude factor, and a
localized spatial envelope. Its strength and spatial localization are controlled by the teacher parameters.

Learned force decoder. The learned model replaces the fixed teacher parameters with an 
generated trajectory. It combines vortex component with low-frequency divergence-free
Fourier modes. 

The flow is driven by an admissible smooth periodic forcing field.
A neural output determines a candidate forcing, which is subsequently
fixed and used as a prescribed input for validation, replay, and
continuum-certification calculations. The admissible forcing family is
smooth and satisfies the prescribed amplitude bound.

\subsection{The discrete Fourier formulation}
\label{sec:fourier_formulation}

Let the uniform grid be
\begin{equation*}
\mathbf x_{\mathbf j}
=\left(\frac{2\pi j_x}{N},\frac{2\pi j_y}{N},\frac{2\pi j_z}{N}\right),
\qquad
j_x,j_y,j_z=0,\ldots,N-1,
\end{equation*}
with
\begin{equation*}
\Delta x=\frac{2\pi}{N}.
\end{equation*}
For even $N$, let $\mathcal K_N$ denote the discrete FFT wave-number set. 

\subsubsection{Fourier transform}
\label{sec:expanded_fourier_convection}

The Fourier equations are compactly by
\begin{equation*}
\partial_t\widehat{\mathbf u}
=\widehat{\mathbf N}
-\nu K^2\widehat{\mathbf u}
+\widehat{\mathbf f}
-i\mathbf k\widehat p.
\end{equation*}

\subsection{Time integration and rollout control}
\label{sec:time_integration_rollout}

The flow trajectories are generated by a stable numerical evolution of the
discretized Navier--Stokes system. Numerical controls are used only to maintain
a consistent and stable rollout over the prescribed time horizon.

\endgroup


\section{Neural Network Architecture}
\label{sec:network}

\begingroup
\setlength{\abovedisplayskip}{4pt plus 1pt minus 1pt}
\setlength{\belowdisplayskip}{4pt plus 1pt minus 1pt}
\setlength{\abovedisplayshortskip}{2pt plus 0.5pt}
\setlength{\belowdisplayshortskip}{2pt plus 0.5pt minus 0.5pt}
\setlength{\jot}{2pt}

The network maps a short history of three-dimensional flow states to a
conditional probability distribution over complete forcing
trajectories.


\subsection{The Physics-Informed observation construction}
\label{sec:detailed_observation}

Each flow snapshot is converted into a compact physics-informed observation
\begin{equation*}
\mathcal{O}_t=(P_t,X_t,S_t,G_t),
\end{equation*}
where $P_t$, $X_t$, $S_t$ and
$G_t$ represent the features. A short history of these
observations is used as the network input. 

The complete forcing architecture is summarized in
Fig.~\ref{fig:sampling_architecture}.

\begin{figure}[!htbp]
\centering
\resizebox{0.96\textwidth}{!}{%
\begin{tikzpicture}[
 x=1cm,y=1cm,
 font=\small,
 >=Latex,
 block/.style={
 draw,
 rounded corners=2mm,
 align=center,
 text width=2.95cm,
 minimum height=0.98cm,
 inner sep=2.3mm
 },
 wideblock/.style={
 draw,
 rounded corners=2mm,
 align=center,
 text width=3.60cm,
 minimum height=1.02cm,
 inner sep=2.35mm
 },
 tinyblock/.style={
 draw,
 rounded corners=1.5mm,
 align=center,
 text width=2.95cm,
 minimum height=0.88cm,
 inner sep=1.9mm
 },
 arrow/.style={->,thick}
]

\node[block, fill=SoftBlueA] (01) at (3.0,16.8) {
\textbf{Velocity}
};
\node[wideblock, fill=SoftBlueA] (02) at (8.3,16.8) {
\textbf{observation}
};
\draw[arrow] (01) -- (02);

\node[wideblock, fill=SoftCyan] (03) at (0.0,13.7) {
\textbf{features}
};
\node[tinyblock, fill=SoftGreenA] (04) at (4.1,13.7) {
\textbf{features}
};
\node[tinyblock, fill=SoftPurple] (05) at (8.2,13.7) {
\textbf{features}
};
\node[tinyblock, fill=SoftBlueB] (06) at (12.3,13.7) {
\textbf{features}
};

\draw[arrow] (02.south) -- ++(0,-0.35) -| (03.north);
\draw[arrow] (02.south) -- ++(0,-0.35) -| (04.north);
\draw[arrow] (02.south) -- ++(0,-0.35) -| (05.north);
\draw[arrow] (02.south) -- ++(0,-0.35) -| (06.north);

\node[wideblock, fill=SoftCyan] (07) at (0.0,10.7) {
\textbf{Encoding}
};
\node[wideblock, fill=SoftRose] (08) at (0.0,8.0) {
\textbf{Encoding}
};
\node[wideblock, fill=SoftRose] (09) at (0.0,5.25) {
\textbf{Encoder}
};
\draw[arrow] (03) -- (07);
\draw[arrow] (07) -- (08);
\draw[arrow] (08) -- (09);

\node[wideblock, fill=SoftGreenA] (10) at (4.1,8.0) {
\textbf{Encoder}
};
\node[wideblock, fill=SoftPurple] (11) at (8.2,8.0) {
\textbf{Encoder}
};
\node[wideblock, fill=SoftBlueB] (12) at (12.3,8.0) {
\textbf{Encoder}
};
\draw[arrow] (04) -- (10);
\draw[arrow] (05) -- (11);
\draw[arrow] (06) -- (12);

\node[wideblock, fill=SoftRose] (13) at (4.1,5.25) {
\textbf{Encoding}
};
\draw[arrow] (09) -- (13);
\draw[arrow] (10) -- (13);

\node[wideblock, fill=SoftCyan] (14) at (8.2,5.25) {
\textbf{Encoding}
};
\draw[arrow] (13) -- (14);
\draw[arrow] (11) -- (14);
\draw[arrow] (12) -- (14);

\node[block, fill=SoftCyan] (15) at (12.3,5.25) {
\textbf{Encoder}
};
\draw[arrow] (14) -- (15);

\node[wideblock, fill=SoftBlueC] (16) at (10.25,2.45) {
\textbf{Encoder}
};
\draw[arrow] (15.south) -- ++(0,-0.35) -| (16.north);

\node[wideblock, fill=SoftPink] (17) at (5.95,2.45) {
\textbf{Encoder}
};
\draw[arrow] (16) -- (17);

\node[wideblock, fill=SoftLavender] (18) at (1.65,2.45) {
\textbf{Encoding}
};
\draw[arrow] (17) -- (18);

\node[wideblock, fill=SoftLavender] (19) at (1.65,-0.45) {
\textbf{Parameters}
};
\draw[arrow] (18) -- (19);

\node[wideblock, fill=SoftPink] (20) at (6.0,-0.45) {
\textbf{Decoder}
};
\draw[arrow] (19) -- (20);

\node[wideblock, fill=SoftGreenB] (21) at (10.55,-0.45) {
\textbf{Navier-Stokes}
};
\draw[arrow] (20) -- (21);

\node[block, fill=SoftBlueD] (22) at (4.0,-3.10) {
\textbf{Vorticity}
};
\node[block, fill=SoftBlueD] (23) at (8.0,-3.10) {
\textbf{Velocity}
};
\node[block, fill=SoftBlueD] (24) at (12.0,-3.10) {
\textbf{Force}
};
\draw[arrow] (21.south) -- ++(0,-0.35) -| (22.north);
\draw[arrow] (21.south) -- ++(0,-0.35) -| (23.north);
\draw[arrow] (21.south) -- ++(0,-0.35) -| (24.north);

\end{tikzpicture}%
}

\caption{The Structure}
\label{fig:sampling_architecture}
\end{figure}
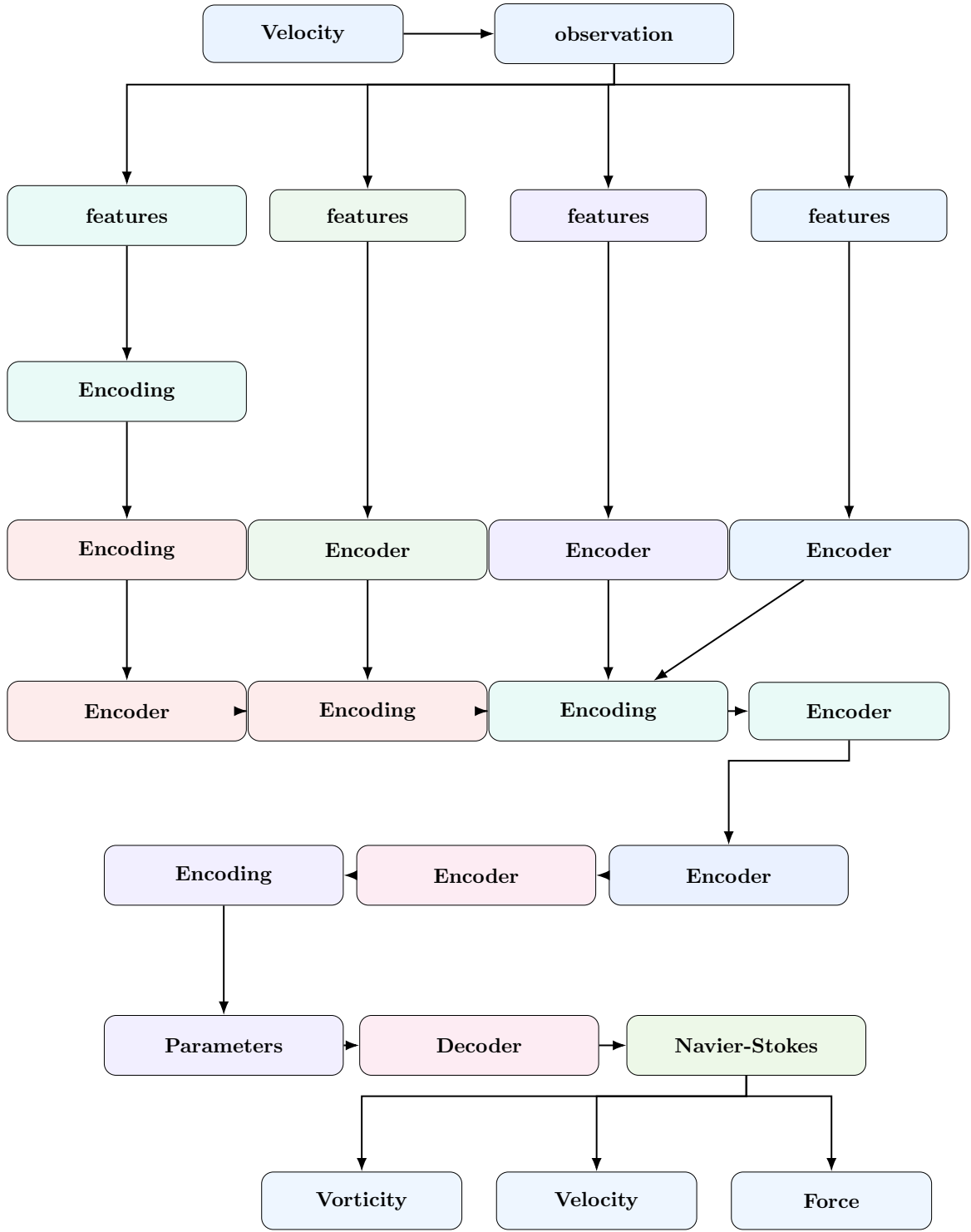

\subsection{Neural feature encoding and generative forcing model}

The neural architecture maps a short history of flow observations into learned features and uses them to generate admissible forcing trajectories. The detailed feature construction, intermediate network components, and internal parameterization are implementation-specific and are not required by the subsequent mathematical analysis.


\section{Forcing Parameterization}
\label{sec:forcing}

The neural output $Y$ parameterizes an admissible external forcing field
through a decoding map
\begin{equation*}
Y
\xrightarrow{\mathcal D}
\mathbf f_Y.
\end{equation*}
The neural forcing specifies
the geometric vortex component, the divergence-free Fourier component, their relative mixing,
and the forcing amplitude.

The decoder is constructed so that the resulting forcing is spatially
periodic, compatible with the incompressible formulation, and smoothly bounded
by a prescribed amplitude scale,
\begin{equation*}
\|\mathbf f_Y(\mathbf x,t)\|_2
\leq f_b,
\qquad
f_b=10^{21}.
\end{equation*}

The finite-dimensional forcing parameters are converted into a smooth
-time field by spatial reconstruction and smooth temporal encoding.
Consequently,
\begin{equation*}
\mathbf f_Y
\in C^\infty(\mathbb T^3\times\mathbb R).
\end{equation*}

Thus the complete forcing construction may be represented abstractly as
\begin{equation*}
Y
\xrightarrow{\mathcal D}
\mathbf f_Y,
\qquad
\mathbf f_Y\in C^\infty,
\qquad
\|\mathbf f_Y\|_\infty\leq f_b.
\end{equation*}

The corresponding
forcing field is fixed and used as a prescribed external input in the
subsequent validation and continuum-certification calculations.


\section{Physical Target Quantities}
\label{sec:targets}

The physical objectives are characterized by the maximum vorticity and its accumulated value over time. These quantities provide the extreme-vorticity diagnostics used in the optimization objectives.


\section{Physics-Based objectives for the PPO-Clip and Differentiable Methods}
\label{sec:physics_objective}

The two optimization strategies use the same physical diagnostics but different
extreme vorticity objectives. Their losses are summarized only at the level of
composition.

\subsection{PPO-Clip Method}

The PPO physical loss combines an extreme-vorticity objective, an integrated-vorticity objective, and an incompressibility penalty.

\subsection{Differentiable Method}

The differentiable physical loss combines an extreme-vorticity objective with an integrated-vorticity objective over the full rollout horizon.


\section{Differentiable Method}
\label{sec:diff}

The differentiable formulation maps the network parameters through the learned forcing construction into the resulting flow response. The network parameters are optimized through the resulting physical objective.

\section{PPO-Clip Method}
\label{sec:ppo}

The PPO is used only to sample admissible forcing trajectories and map them to bounded forcing inputs.

\section{Fixed-Force $C^\infty$ Replay Validation}
\label{sec:replay}

For validation, a selected trajectory $Y^*$ is decoded into a smooth forcing
$\mathbf f_{\mathrm{ref}}$ and then held fixed during an independent
Navier--Stokes replay. The validation pathway is
\begin{equation*}
Y^*\longrightarrow \mathbf f_{\mathrm{ref}}^{C^\infty}
\longrightarrow \mathrm{NSE}
\longrightarrow \mathbf U_{\mathrm{replay}}(t).
\end{equation*}
The replay therefore tests the response to the prescribed inferred forcing
without further neural feedback.

\section{Experimental and Validation Method}
\label{sec:experimental}

Selected forcing trajectories are validated by direct Navier--Stokes solves and independent fixed-force replay beyond the training horizon.


\section{Results and Physical Interpretation}
\label{sec:results}

Figure~\ref{fig:result_two_routes} shows one representative solver-validated
trajectory from each optimization strategy. Both examples exhibit strong
vorticity amplification and are independently checked by fixed-force
Navier--Stokes replay. They are illustrative trajectories rather than a
statistical characterization of all generated candidates.

\begin{figure}[!htbp]
\centering
\begin{minipage}[t]{0.485\textwidth}
\centering
\safeincludegraphics[width=\linewidth]{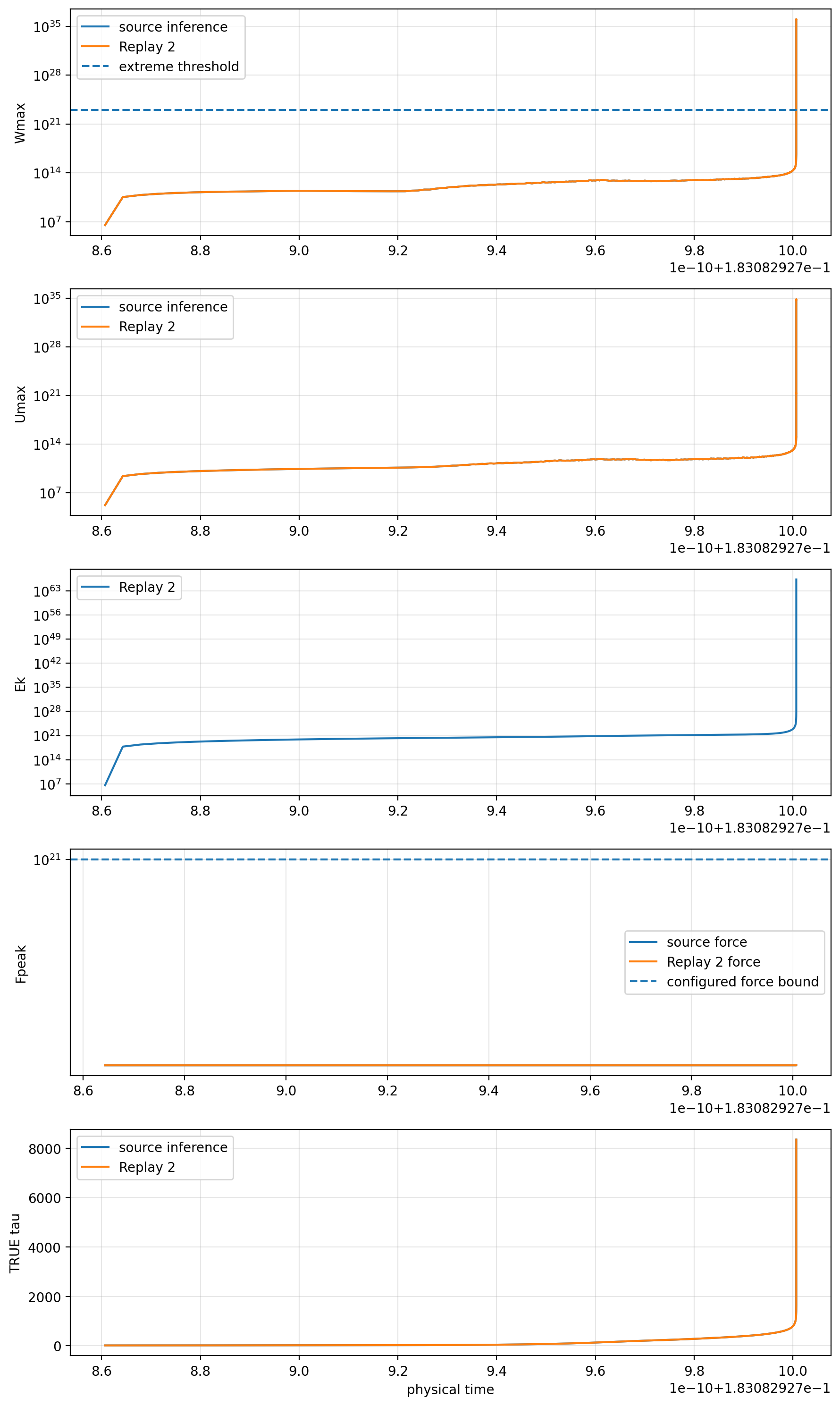}
\smallskip
\textbf{(a) Differentiable}
\end{minipage}\hfill
\begin{minipage}[t]{0.485\textwidth}
\centering
\safeincludegraphics[width=\linewidth]{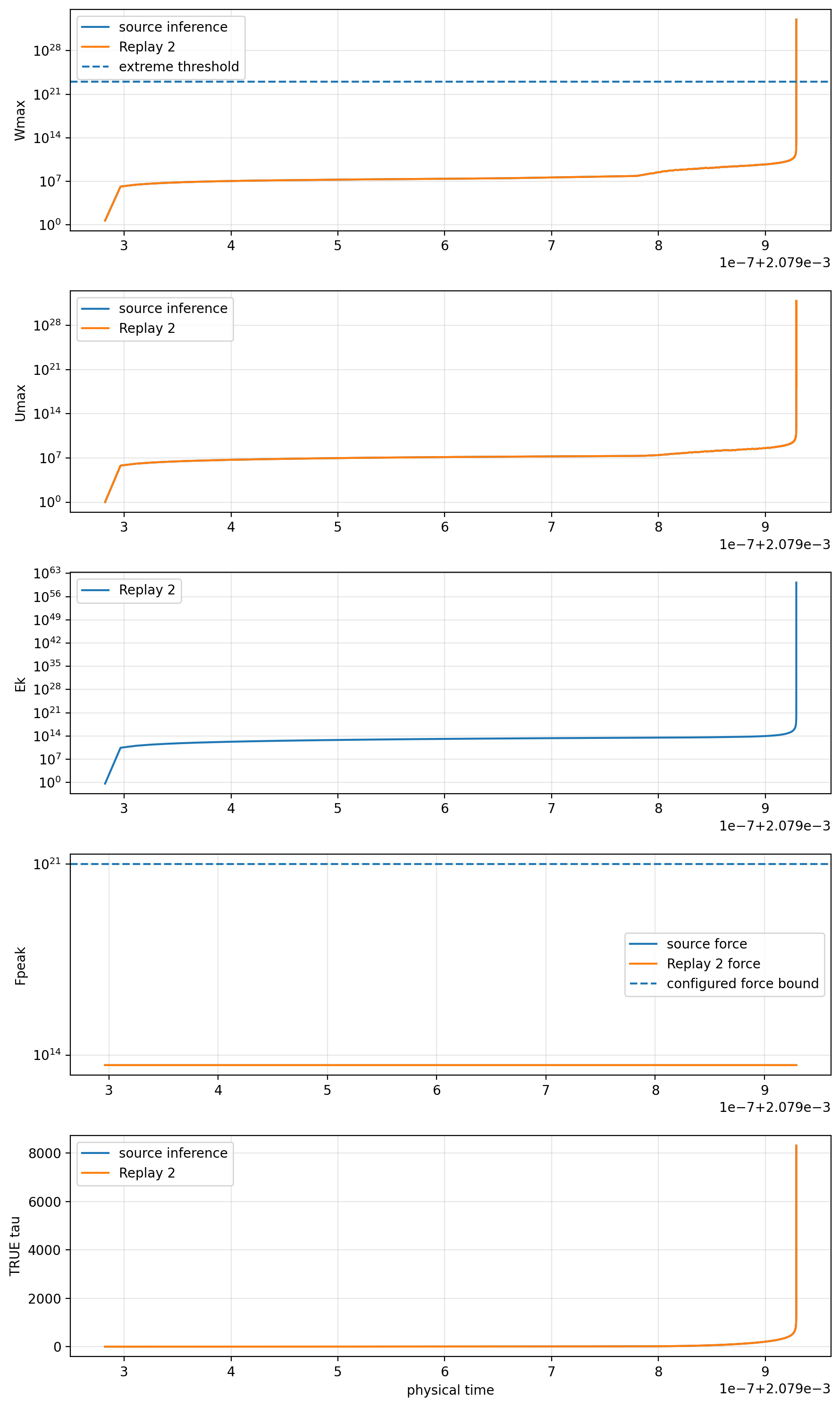}
\smallskip
\textbf{(b) PPO-Clip}
\end{minipage}
\caption{The solver-validated trajectories generated by the two
optimization strategies.}
\label{fig:result_two_routes}
\end{figure}

\endgroup


\section{Conclusion}
\label{sec:conclusion}

The framework identifies admissible forcing trajectories from flow data and validates selected candidates by independent fixed-force Navier--Stokes replay. Continuum singularity claims are treated separately by the mathematical certification in Part~II.


\part{Continuum Blow-Up Certification}

\setcounter{section}{0}
\setcounter{subsection}{0}
\setcounter{subsubsection}{0}

\noindent\textbf{Logical role of Part II.}
Part~I constructs and freezes neural forcing candidates and provides independent finite-resolution replay diagnostics. Part~II does not treat computational growth as a continuum singularity proof. Instead, it supplies the deterministic certification and probability arguments needed to turn a rigorously established continuum reciprocal-vorticity certificate into a conditional finite-time blow-up statement.

\section{Proof Architecture and Scope}

The purpose of this work is to make precise a neural strategy to finite-time blow-up for forced three-dimensional incompressible Navier--Stokes flow without conflating computational explosion, optimizer behavior, or finite-resolution growth with a continuum singularity proof.

The central methodological principle is a strict separation between neural search and mathematical certification. The neural model is used only to identify a forcing candidate. After a candidate has been selected, the forcing is frozen and the continuum PDE argument proceeds independently of the network parameters.

The proof is organized into two complementary strategies. Strategy~I is a validated certificate-construction strategy. Rigorously corrected finite-resolution information is transferred to a continuum reciprocal-vorticity certificate. Strategy~II is a continuum probability-closure strategy. Starting from a strict continuum certificate supplied either by Strategy~I or by an independent continuum loss theorem, the certificate is propagated to a robust open parameter set and then combined with nondegeneracy of the neural output law to obtain positive probability of finite-time blow-up.

The deterministic mechanism is an integrated reciprocal-vorticity condition. It requires a persistent decrease of reciprocal maximum vorticity from a chosen certificate-start time and yields Riccati-type growth of maximum vorticity. This formulation is weaker than requiring positivity of instantaneous normalized production at every time and is compatible with temporary negative stretching. The certificate-start time is a later reference point along the already evolved trajectory; it is not a reset of the Navier--Stokes initial condition.

Direct recomputation from the stored trajectories and independent fixed-forcing replays gives positive finite-resolution reciprocal-vorticity growth for the archived neural candidates. In the late persistent regime, the available trajectories approach their corresponding Riccati comparison times extremely closely. The continuum implication and probability closure are proved in Part~II. The archived Differentiable and PPO-Clip trajectories require no further analytical argument. Verification of this condition allows Strategy~I to certify the corresponding trajectory.

\paragraph{Related literature.}
The mathematical regularity theory of the three-dimensional incompressible
Navier--Stokes equations provides the analytical background for the present
work. Standard references include the partial-regularity theory of
Caffarelli, Kohn, and Nirenberg and the critical regularity result of
Escauriaza, Seregin, and \v{S}ver\'ak
\cite{CaffarelliKohnNirenberg1982,EscauriazaSereginSverak2003}. The use of maximum vorticity and
reciprocal-vorticity growth here is also conceptually related to the
vorticity-based blow-up philosophy of Beale, Kato, and Majda for the
three-dimensional Euler equations \cite{beale1984bkm}. The present argument is not
an application of the Beale--Kato--Majda theorem. It concerns a forced viscous
Navier--Stokes system and uses an integrated Riccati certificate tailored to
that system.

A second relevant line of work concerns rigorous computational certification for
Navier--Stokes equations. Computer-assisted verification methods have also
been developed for stationary three-dimensional Navier--Stokes problems using
explicit quantitative error bounds; see, for example, Liu, Nakao, and Oishi
\cite{LiuNakaoOishi2022}. Strategy~I is closest in spirit to this validated-
computation literature, but it is tailored to reciprocal-vorticity and
production-ratio certificates for a single frozen neural forcing.

Machine-learning approaches to PDEs have meanwhile developed operator-learning
architectures such as the Fourier Neural Operator
\cite{LiEtAl2021}. In contrast, the neural component here is not
used as a substitute for a continuum PDE solver and computational growth is not
used as proof of regularity blow-up. The network is used only to generate a
finite-dimensional forcing candidate; every singularity conclusion is then
obtained independently from a continuum certificate, either directly or
through validated computational transfer.

The proof proceeds in two stages. First, a frozen neural forcing candidate is connected to a deterministic continuum certificate, either through validated computational transfer or through a direct continuum argument. Second, the certificate is propagated through robustness and the integrated Riccati mechanism to finite-time loss of smooth continuation. When the neural output law assigns positive mass to a certified neighborhood, the same argument yields a positive-probability conclusion.

\section{Neural forcing family}

The computational pipeline is represented schematically by
\begin{eqnarray*}
\mathcal N_\theta
=
\mathcal D\circ\mathcal M_\theta\circ\mathcal T_\theta\circ\mathcal E_\theta,
\end{eqnarray*}
where an encoding stage and a temporal module process a four-frame flow history, a stochastic module produces a forcing trajectory, and the final mapping sends that trajectory into a physical forcing.

Let $\mathcal Y$ denote a finite-dimensional admissible forcing-parameter set. The neural output is represented abstractly by
\begin{eqnarray*}
Y\in\mathcal Y.
\end{eqnarray*}
For a fixed history $H$, the network induces a conditional law
\[
Y\sim\mu_\theta(\cdot\mid H).
\]
Deterministic certification requires only that the network identify a candidate $Y^*$. We then freeze
\begin{eqnarray*}
f^*(x,t)=\mathcal D(Y^*)(x,t).
\end{eqnarray*}

The forcing is represented on a fixed finite Fourier lattice $\Lambda_f\subset\mathbb Z^3\setminus\{0\}$ and reconstructed in time by normalized Gaussian radial-basis weights. Hence
\begin{eqnarray*}
f^*\in C^\infty(\mathbb T^3\times I),
\qquad
\nabla\cdot f^*=0.
\end{eqnarray*}
The decoder also enforces the prescribed pointwise forcing cap, with amplitude bounded by $10^{21}$.

When the PDE resolution is fine enough to contain $\Lambda_f$, the forcing is resolution-independent.
\begin{eqnarray*}
P_Nf^*=f^*,\qquad N\ge N_f.
\end{eqnarray*}
Thus refinement changes only the PDE approximation, not the physical forcing under test.

\subsection{Direct continuum interpretation of the frozen decoder}

The continuum forcing is defined directly from the retained neural Fourier
coefficients; it is not defined as a limit of computational PDE solutions. At the
forcing knots $j=0,\ldots,n$, write
\begin{eqnarray*}
 f_j^Y(x)
 =
 \sum_{k\in\mathcal K}\widehat f_{j,k}^{\,Y}e^{ik\cdot x},
 \qquad |\mathcal K|<\infty,
\end{eqnarray*}
and reconstruct in time with normalized Gaussian-RBF weights,
\begin{eqnarray*}
 f_Y(x,t)=\sum_{j=0}^{n}\alpha_j(t)f_j^Y(x),
 \qquad
 \alpha_j(t)>0,
 \qquad
 \sum_{j=0}^{n}\alpha_j(t)=1.
\end{eqnarray*}
Thus $f_Y$ is an exact smooth trigonometric polynomial with smooth
time dependence. No computational $N\to\infty$ passage is required to define the
continuum forcing family $Y\mapsto f_Y$. Computational refinement is relevant only
when one elects to validate a candidate through a separate computer-assisted
refinement strategy.

\section{Continuum forced Navier--Stokes dynamics}

We consider
\begin{eqnarray*}
\partial_tu+(u\cdot\nabla)u
=-\nabla p+\nu\Delta u+f^*(x,t),
\qquad
\nabla\cdot u=0,
\end{eqnarray*}
on the three-dimensional torus $\mathbb T^3$ with $\nu>0$.

Define
\[
\omega=\nabla\times u,
\qquad
W(t)=\|\omega(t)\|_{L^\infty},
\qquad
S=\frac12(\nabla u+\nabla u^T).
\]
The vorticity equation is
\begin{eqnarray*}
\partial_t\omega+(u\cdot\nabla)\omega
=(\omega\cdot\nabla)u+\nu\Delta\omega+\nabla\times f^*.
\end{eqnarray*}
Since the antisymmetric part of $\nabla u$ contributes zero to the quadratic form,
\[
\omega\cdot(\omega\cdot\nabla)u=\omega\cdot S\omega.
\]
Therefore
\begin{eqnarray*}
\frac12(\partial_t+u\cdot\nabla)|\omega|^2
=
\omega\cdot S\omega
+\nu\omega\cdot\Delta\omega
+\omega\cdot\nabla\times f^*.
\end{eqnarray*}
Define the total vorticity production
\begin{eqnarray*}
\Pi(x,t)
=
\omega\cdot S\omega
+\nu\omega\cdot\Delta\omega
+\omega\cdot\nabla\times f^*.
\end{eqnarray*}

At an active maximum-vorticity point, let
$e=\omega/W$. Then
\[
\omega\cdot S\omega=W^2e^TSe,
\]
and the dissipative and forcing terms satisfy the conservative bounds
\[
\nu\omega\cdot\Delta\omega
\ge -\nu W\|\Delta\omega\|_\infty,
\qquad
\omega\cdot\nabla\times f^*
\ge -W\|\nabla\times f^*\|_\infty.
\]
Consequently,
\begin{eqnarray*}
\frac{\Pi}{W^3}
\ge
\frac{e^TSe}{W}
-\nu\frac{\|\Delta\omega\|_\infty}{W^2}
-\frac{\|\nabla\times f^*\|_\infty}{W^2}.
\end{eqnarray*}
This pointwise lower estimate is useful diagnostically, but the integrated
Riccati strategy developed below does not require its right-hand side to remain
positive at every instant.

Let
\[
\mathcal M(t)=\{x:|\omega(x,t)|=W(t)\}.
\]
At an active maximum point the spatial gradient of $|\omega|^2$ vanishes, so the transport term drops out of the maximum-envelope calculation.

\section{Pointwise continuum Riccati certificate}

Define
\begin{eqnarray*}
\mathscr C(f^*;I)
=
\inf_{\substack{t\in I\\x\in\mathcal M(t)}}
\frac{\Pi(x,t)}{W(t)^3}.
\end{eqnarray*}
Assume $W(t)>0$ on $I$ and
\begin{eqnarray*}
\mathscr C(f^*;I)\ge c_0>0.
\end{eqnarray*}
Then at every active maximum point
\[
\Pi(x,t)\ge c_0W(t)^3.
\]
Setting $M(t)=W(t)^2$, the Dini-envelope argument gives
\[
D_+M(t)\ge2c_0W(t)^3.
\]
Since $M=W^2$,
\begin{eqnarray*}
D_+W(t)\ge c_0W(t)^2.
\end{eqnarray*}
This may be written almost everywhere over smooth compact subintervals as
\[
W'(t)\ge c_0W(t)^2.
\]
Hence
\[
\left(\frac1W\right)'\le-c_0,
\]
and integration yields
\begin{eqnarray*}
\frac1{W(t)}
\le
\frac1{W(t_0)}-c_0(t-t_0).
\end{eqnarray*}
Equivalently,
\begin{eqnarray*}
W(t)
\ge
\frac{W_0}{1-c_0W_0(t-t_0)},
\qquad
W_0=W(t_0).
\end{eqnarray*}
The Riccati comparison time is
\begin{eqnarray*}
T_R=t_0+\frac1{c_0W_0}.
\end{eqnarray*}
If the certificate persists on $[t_0,T_R)$, a bounded classical solution cannot continue smoothly through $T_R$.

At a maximum point,
\[
\omega\cdot\Delta\omega
=\frac12\Delta|\omega|^2-|\nabla\omega|^2\le0,
\]
so viscosity remains fully inside the certificate. The strict cubic condition is precisely that stretching and forcing production overcome viscous loss by a positive $W^3$ margin.

\section{Integrated reciprocal-vorticity certificate}

The neural trajectories motivate the weaker integrated certificate because instantaneous stretching may be negative at isolated times.

Assume $W_Y(t)>0$ on the classical interval and define the reciprocal variable
\begin{eqnarray*}
q_Y(t)=\frac1{W_Y(t)}.
\end{eqnarray*}
The reciprocal variable $q_Y$ is locally absolutely continuous on smooth compact subintervals. At
almost every $t$, define
\begin{eqnarray*}
 a_Y(t)=\frac{W_Y'(t)}{W_Y(t)^2}=-q_Y'(t).
\end{eqnarray*}
Then
\begin{eqnarray*}
\frac1{W_Y(t)}
=
\frac1{W_Y(t_0)}-\int_{t_0}^{t}a_Y(s)\,ds.
\end{eqnarray*}
Thus the fixed-start reciprocal-vorticity coefficient is the time average
\begin{eqnarray*}
\frac{W_Y(t_0)^{-1}-W_Y(t)^{-1}}{t-t_0}
=
\frac1{t-t_0}\int_{t_0}^{t}a_Y(s)\,ds.
\end{eqnarray*}
This identity makes explicit why temporary intervals with $a_Y<0$, and hence
local negative maximum-vorticity growth or negative stretching, are compatible
with a positive cumulative certificate.

Fix a starting time $t_0$ and define
\begin{eqnarray}
\mathcal C_T(Y)
=
\inf_{t\in(t_0,T]}
\frac{W_Y(t_0)^{-1}-W_Y(t)^{-1}}{t-t_0}.
\label{eq:CT}
\end{eqnarray}
If
\[
\mathcal C_T(Y)\ge c_0>0,
\]
then
\begin{eqnarray*}
\frac1{W_Y(t)}
\le
\frac1{W_Y(t_0)}-c_0(t-t_0),
\end{eqnarray*}
and therefore
\begin{eqnarray*}
W_Y(t)
\ge
\frac{W_Y(t_0)}{1-c_0W_Y(t_0)(t-t_0)}.
\end{eqnarray*}
Thus the classical solution cannot extend smoothly through
\begin{eqnarray*}
T_R(Y)=t_0+\frac1{c_0W_Y(t_0)}.
\end{eqnarray*}

An even weaker sequence formulation suffices. Suppose there exists $t_j\uparrow T_R$ such that
\begin{eqnarray*}
\frac1{W(t_j)}
\le
\frac1{W(t_0)}-c_0(t_j-t_0).
\end{eqnarray*}
Since $W(t_0)^{-1}=c_0(T_R-t_0)$,
\[
\frac1{W(t_j)}\le c_0(T_R-t_j),
\]
and hence
\begin{eqnarray*}
W(t_j)\ge\frac1{c_0(T_R-t_j)}\longrightarrow+\infty.
\end{eqnarray*}
Therefore smooth continuation through $T_R$ is impossible. Pointwise positivity of $\Pi/W^3$ at every instant is not required.

\paragraph{Optional computational validation.}
Strategy~I collects the finite-resolution certificates, the validated error enclosures used for transfer to the continuum, and the refinement theorem. This material provides an independent computer-assisted validation strategy and is not a logical prerequisite of the direct continuum proof developed below.

\section{Existence of the Neural Loss Minimizer and Riccati-Certificate Feasibility}
\label{sec:loss-minimizer-feasibility}

This section isolates the neural optimization statement from the PDE certificate statement. The two are related, but they are not the same theorem. The first result is existence of a minimizer of the finite-resolution neural objective; the second is the exact identification of the zero level set of a Riccati-certificate loss.

\subsection{Finite-dimensional neural-to-loss map}

At fixed spectral resolution $N$, finite rollout horizon, and a fixed auxiliary sampling variable, write the complete map schematically as
\begin{eqnarray*}
\theta
\xrightarrow{\mathcal E}
 h_\theta
\xrightarrow{\mathcal T}
 z_\theta
\xrightarrow{\mathcal G}
 Y_\theta
\xrightarrow{\mathcal D}
 f_\theta
\xrightarrow{\Phi_N}
 u_{\theta,N}
\xrightarrow{\mathcal Q}
 \mathcal L_N(\theta).
\end{eqnarray*}
Hence
\begin{eqnarray*}
\mathcal L_N:\Theta\to\mathbb R.
\end{eqnarray*}
At fixed grid size the deterministic neural layers are finite-dimensional compositions of continuous maps. The stochastic module is assumed to admit a continuous realization for each fixed auxiliary sampling variable on the admissible parameter set. Consequently,
\begin{eqnarray*}
\theta\mapsto Y_\theta\quad\text{continuous}.
\end{eqnarray*}

\subsection{Continuity and boundedness of the forcing decoder}

The decoder uses the smooth radial saturation
\begin{eqnarray*}
S_b(z)
=
\frac{z}{\sqrt{1+\|z\|^2/b^2}},
\qquad
b=10^{21}.
\end{eqnarray*}
Consequently the decoded force is pointwise bounded by the prescribed amplitude scale,
\begin{eqnarray*}
\|f_Y\|<10^{21},
\end{eqnarray*}
and $Y\mapsto f_Y$ is continuous. The spatial forcing is a finite Fourier polynomial,
\begin{eqnarray*}
f_Y(x,t)
=
\sum_{k\in\Lambda_f}
\widehat f_k(Y,t)e^{ik\cdot x},
\qquad |\Lambda_f|<\infty,
\end{eqnarray*}
while the Leray projection and normalized Gaussian-RBF time interpolation are continuous linear or smooth finite-dimensional operations. Thus the complete forcing map
\begin{eqnarray*}
\mathcal F:Y\mapsto f_Y
\end{eqnarray*}
is well defined, continuous, and bounded on bounded parameter sets.

\subsection{Global fixed-$N$ Fourier rollout and finiteness of the loss}

At fixed $N$ the spectral PDE is a finite-dimensional ODE of the form
\begin{eqnarray*}
\dot u_N
=
-\nu A_Nu_N
+
B_N(u_N,u_N)
+
P_Nf_Y.
\end{eqnarray*}
Let $u_N^\circ$ denote the mean-zero component. The standard energy cancellation for the incompressible quadratic term gives
\begin{eqnarray*}
\frac12\frac{d}{dt}\|u_N^\circ\|_2^2
+
\nu\|\nabla u_N^\circ\|_2^2
=
(f_Y,u_N^\circ).
\end{eqnarray*}
Using Poincare and Young inequalities,
\begin{eqnarray*}
\frac{d}{dt}\|u_N^\circ\|_2^2
+
\nu\lambda_1\|u_N^\circ\|_2^2
\le
\frac1{\nu\lambda_1}\|f_Y\|_2^2.
\end{eqnarray*}
Since $\|f_Y\|_2\le C_{\mathbb T^3}10^{21}$, Gronwall yields
\begin{eqnarray*}
\|u_N^\circ(t)\|_2^2
\le
 e^{-\nu\lambda_1(t-t_0)}
 \|u_N^\circ(t_0)\|_2^2
+
\frac{C_{\mathbb T^3}^2\,10^{42}}{\nu^2\lambda_1^2}.
\end{eqnarray*}
Thus the finite-dimensional Fourier trajectory exists globally on every finite time interval. In particular all norms are equivalent at fixed $N$, so
\begin{eqnarray*}
W_N(t)=\|\omega_N(t)\|_\infty<\infty
\end{eqnarray*}
for finite $t$, and therefore
\begin{eqnarray*}
\tau_N(T)
=
\int_{t_0}^T W_N(t)\,dt
<\infty.
\end{eqnarray*}
Hence all finite-$N$ vorticity, accumulated-vorticity, divergence, entropy, and regularization terms entering the training objective are well defined. In particular,
\begin{eqnarray*}
-\infty<\mathcal L_N(\theta)<+\infty.
\end{eqnarray*}
This also records an important logical boundary. A fixed-$N$ Fourier system with bounded forcing does not itself develop a genuine finite-time ODE singularity.

\subsection{Continuity of the neural loss}

Finite-dimensional ODE solutions depend continuously on initial data, forcing, and parameters over finite time intervals. Therefore
\begin{eqnarray*}
\mathcal L_N
=
\mathcal Q
\circ\Phi_N
\circ\mathcal F
\circ\mathcal R_\varepsilon
\circ\mathcal M
\circ\mathcal T
\circ\mathcal E.
\end{eqnarray*}
Every factor in this composition is continuous, and hence
\begin{eqnarray*}
\theta_n\to\theta
\quad\Longrightarrow\quad
\mathcal L_N(\theta_n)\to\mathcal L_N(\theta).
\end{eqnarray*}
Equivalently,
\begin{eqnarray*}
\mathcal L_N\in C(\Theta).
\end{eqnarray*}

\subsection{Existence of a neural loss minimizer}

Restrict the parameters to the compact ball
\begin{eqnarray*}
\Theta_R
=
\{\theta\in\mathbb R^P:\|\theta\|\le R\},
\end{eqnarray*}
continuity and the Weierstrass theorem give
\begin{eqnarray*}
\exists\,\theta_N^*\in\Theta_R:
\qquad
\mathcal L_N(\theta_N^*)
=
\min_{\theta\in\Theta_R}\mathcal L_N(\theta).
\end{eqnarray*}
Thus a finite-$N$ neural loss minimizer exists on every compact parameter domain.

An alternative is to optimize over all of $\mathbb R^P$ after adding coercive parameter regularization,
\begin{eqnarray*}
\widetilde{\mathcal L}_N(\theta)
=
\mathcal L_N(\theta)
+
\lambda_\theta\|\theta\|_2^2,
\qquad
\lambda_\theta>0.
\end{eqnarray*}
If $\mathcal L_N(\theta)\ge-C_N$, then
\begin{eqnarray*}
\widetilde{\mathcal L}_N(\theta)
\ge
-C_N+\lambda_\theta\|\theta\|^2
\longrightarrow+\infty
\qquad
(\|\theta\|\to\infty).
\end{eqnarray*}
Hence $\widetilde{\mathcal L}_N$ is continuous and coercive, so
\begin{eqnarray*}
\exists\,\theta_N^*\in\mathbb R^P:
\qquad
\widetilde{\mathcal L}_N(\theta_N^*)
=
\min_{\theta\in\mathbb R^P}\widetilde{\mathcal L}_N(\theta).
\end{eqnarray*}
This is an existence statement for a minimizer; it is not a theorem that a particular stochastic optimizer such as Adam or PPO converges to that minimizer.

\subsection{Riccati certificate loss and its exact zero level set}

Let $\mathcal M_N(t)$ denote the active computational maximum-vorticity set and define the finite-resolution normalized production certificate
\begin{eqnarray*}
\mathscr C_N(Y)
=
\inf_{\substack{t\in I\\x\in\mathcal M_N(t)}}
\frac{
\omega_N\cdot S_N\omega_N
+
\nu\omega_N\cdot\Delta\omega_N
+
\omega_N\cdot\nabla\times f_Y
}{W_N(t)^3}.
\end{eqnarray*}
For a prescribed target $c_{\rm target}>0$, define
\begin{eqnarray*}
L_{\rm Ric,N}(Y)
=
[c_{\rm target}-\mathscr C_N(Y)]_+^2.
\end{eqnarray*}
Because $[a]_+=0$ if and only if $a\le0$,
\begin{eqnarray*}
L_{\rm Ric,N}(Y)=0
\quad\Longleftrightarrow\quad
\mathscr C_N(Y)\ge c_{\rm target}.
\end{eqnarray*}
Equivalently, the zero level set is precisely the finite-resolution cubic-production certificate set,
\begin{eqnarray*}
\omega_N\cdot S_N\omega_N
+
\nu\omega_N\cdot\Delta\omega_N
+
\omega_N\cdot\nabla\times f_Y
\ge
c_{\rm target}W_N^3
\end{eqnarray*}
throughout the certified interval and at every active maximum-vorticity point. Thus the certificate loss is not merely a heuristic reward for large vorticity; its zero set is exactly the stated finite-$N$ Riccati certificate set.

For a finite collection of generated candidates $\{Y_m\}$, define
\begin{eqnarray*}
L_{{\rm Ric},m}
=
[c_{\rm target}-\mathscr C_N(Y_m)]_+^2.
\end{eqnarray*}
These candidate losses are combined by a smooth finite aggregation, denoted by
\begin{eqnarray*}
L_{\rm Ric,gen}
=
\mathcal A\!\left(\{L_{{\rm Ric},m}\}_m\right),
\end{eqnarray*}
where only continuity and boundedness from below are used in the argument below. A corresponding composite objective is
\begin{eqnarray}
L_{\rm total}
=
L_{\rm base}
+
\lambda_RL_{\rm Ric,gen}
+
\lambda_\theta\|\theta\|_2^2.
\label{eq:lm-total}
\end{eqnarray}
If the noncoercive part is continuous and bounded below, then $L_{\rm total}$ is continuous and coercive and therefore attains a global minimum.

\subsection{Minimizer existence versus certificate feasibility}

Two statements must be kept separate. The first is
\begin{eqnarray*}
\exists\,\theta_N^*:
\quad
L_{\rm total}(\theta_N^*)
=
\min_\theta L_{\rm total}(\theta),
\end{eqnarray*}
which follows from continuity plus compactness or coercivity. The second is the certificate feasibility statement
\begin{eqnarray}
\exists\,Y:\qquad
\mathscr C_N(Y)\ge c_{\rm target}.
\label{eq:lm-feasible}
\end{eqnarray}
If \eqref{eq:lm-feasible} is independently established, then the pure certificate loss satisfies
\begin{eqnarray*}
\min_Y[c_{\rm target}-\mathscr C_N(Y)]_+^2=0.
\end{eqnarray*}
If that minimum is attained, there exists $Y_N^*$ with
\begin{eqnarray*}
\mathscr C_N(Y_N^*)\ge c_{\rm target}.
\end{eqnarray*}
Thus, for the pure certificate loss,
\begin{eqnarray*}
\text{feasibility}+\text{attainment}
\Longrightarrow
\text{zero certificate-loss minimizer}.
\end{eqnarray*}
Under a combined objective such as \eqref{eq:lm-total}, feasibility of $L_{\rm Ric,N}=0$ does not by itself imply that every global minimizer has zero certificate loss, because the base objective can trade off against the certificate term. An exact implication requires a constrained formulation, a lexicographic rule, or a proved exact-penalty theorem.

\section{Strategy I Validated Certificate Construction}
\label{sec:refinement}

This section gives the complete refinement-validation proof in the main text because it can supply, rather than merely accompany, the deterministic continuum certificate used by the direct probability strategy. The logical role of Strategy~I is
\[
\begin{array}{c}
\text{finite-resolution neural trajectory}
\\[1mm]
\Downarrow
\\[-1mm]
\text{rigorously error-corrected certificate}
\\[1mm]
\Downarrow
\\[-1mm]
\text{continuum integrated/Riccati certificate}.
\end{array}
\]
Certification of the last implication with a strictly positive residual margin allows Strategy~II to take that continuum certificate as its deterministic input and convert robustness of the certificate into positive blow-up probability. Thus Strategy~I is a complementary certificate-construction mechanism for Strategy~II, not an appendix-only side calculation. 

\subsection{Finite-resolution neural certificate}

Let
\[
W_N(t)=\|\omega_N(t)\|_\infty
\]
be the maximum vorticity of a computational trajectory for the same frozen forcing.

\paragraph{PDE initial condition versus certificate start.}
The Navier--Stokes evolution has its original PDE initial condition
\begin{eqnarray*}
u_N(x,t_{\rm init})=u_{\rm init}(x),
\qquad
t_{\rm init}=0.
\end{eqnarray*}
This initial condition is never reset in the certificate calculation. The
trajectory is evolved continuously from $u_{\rm init}$ through all earlier time
steps. Independently of the PDE initial time, choose a later saved index
$n_c$ at which the Riccati certificate test begins, and define
\begin{eqnarray*}
t_c=t_{n_c},
\qquad
W_c=W_N(t_c).
\end{eqnarray*}
Thus $n_c$ is the \emph{certificate-start index}, not an initial-condition
index, and $t_c$ is the \emph{certificate-start time}, not the PDE initial
time. Schematically,
\begin{eqnarray*}
u_{\rm init}
\longrightarrow
u_N(t_1)
\longrightarrow\cdots\longrightarrow
u_N(t_c)
\longrightarrow\cdots,
\end{eqnarray*}
where the certificate is tested only on the tail beginning at $t_c$.

At every later saved time $t_m>t_c$, equivalently $m>n_c$, define the
fixed-start prefix coefficient
\begin{eqnarray*}
c_{N,m}^{\rm prefix}
=
\frac{W_c^{-1}-W_N(t_m)^{-1}}{t_m-t_c}.
\end{eqnarray*}
Set
\begin{eqnarray*}
c_N^{\rm data}
=
\inf_{m>n_c}c_{N,m}^{\rm prefix}.
\end{eqnarray*}
If $c_N^{\rm data}>0$, every certified prefix obeys
\begin{eqnarray*}
\frac1{W_N(t_m)}
\le
\frac1{W_c}-c_N^{\rm data}(t_m-t_c),
\end{eqnarray*}
and hence
\begin{eqnarray*}
W_N(t_m)
\ge
\frac{W_c}{1-c_N^{\rm data}W_c(t_m-t_c)}.
\end{eqnarray*}
This is a finite-resolution integrated Riccati certificate on the tail
$[t_c,\,t_m]$. It does not change the PDE initial condition and is not by
itself a continuum singularity proof. In the later abstract continuum
Riccati theorem, the reference time denoted there by $t_0$ plays the same
logical role as the certificate-start time $t_c$ used here.

\subsection{The observed neural trajectories}

Using the predefined crossing $W\approx10^{23}$ as the starting point, the two neural trajectories in the manuscript give
\begin{eqnarray}
c_{\rm data}^{(\mathrm{Differentiable})}\approx1.6491136242\times10^{-3},
\label{eq:obs1}
\end{eqnarray}
and
\begin{eqnarray*}
c_{\rm data}^{(\mathrm{PPO\text{-}Clip})}\approx6.4045602779\times10^{-3}.
\end{eqnarray*}
The PPO-Clip trajectory has the dimensionless endpoint quantity
\[
s_{\rm end}=c_{\rm data}W_cT_{\rm data}
\]
is approximately
\begin{eqnarray*}
s_{\rm end}=0.9641427242.
\end{eqnarray*}
A later onset of the strongest persistent regime gives
\begin{eqnarray*}
c_*^{(\mathrm{Differentiable})}\approx6.9473591718\times10^{-3},
\qquad
s_{\rm end}^{(\mathrm{Differentiable})}\approx0.9999963083,
\end{eqnarray*}
and
\begin{eqnarray*}
c_*^{(\mathrm{PPO\text{-}Clip})}\approx6.9663667671\times10^{-3},
\qquad
s_{\rm end}^{(\mathrm{PPO\text{-}Clip})}\approx0.999999998521792.
\end{eqnarray*}
Thus both late trajectories exhibit the common empirical scale
\begin{eqnarray}
c_{\rm neural}\approx6.95\times10^{-3}.
\label{eq:cneural}
\end{eqnarray}
These are finite-resolution data-derived coefficients and are not yet continuum constants.

\subsection{Direct NPZ recomputation and independent replay cross-check}

The finite-resolution coefficients can be recomputed directly from the archived
neural NPZ files and their independent fixed-force replay files. At the extreme
late-time scales considered here, increments in the stored absolute time can fall
below the representable increment of the large base time. Therefore, for this
cross-check, the physical prefix duration is reconstructed from the saved replay
timesteps,
\begin{eqnarray*}
T_{n_c,m}
=
\sum_{k=n_c+1}^{m}\Delta t_k,
\end{eqnarray*}
and the prefix coefficient is evaluated as
\begin{eqnarray*}
c_{n_c,m}^{\rm replay}
=
\frac{W_{n_c}^{-1}-W_m^{-1}}{T_{n_c,m}}.
\end{eqnarray*}

The Differentiable trajectory uses the certificate-start index $n_c=4695$, chosen at the
predefined crossing near $W\approx10^{23}$. This does not mean that the PDE is
initialized at step $4695$; the state at that step is reached by evolving from
$u_{\rm init}$ at $t=0$. The smallest
fixed-start prefix coefficient occurs at $m=4707$, for which
\begin{eqnarray*}
W_{n_c}&=&1.0058517286\times10^{23},
\nonumber\\
W_m&=&1.0220864799\times10^{23},
\nonumber\\
T_{n_c,m}&=&9.5757643698\times10^{-23}.
\end{eqnarray*}
Consequently,
\begin{eqnarray}
c_{\rm data,NPZ}^{(\mathrm{Differentiable})}
=
1.6491136240\times10^{-3}.
\label{eq:npz-Differentiable-cdata}
\end{eqnarray}
The absolute differences between the original neural trajectory and the
independent replay at these two endpoints are
\begin{eqnarray*}
e_{W,a}^{\rm proxy}
&=&1.1249794417\times10^{12},
\nonumber\\
e_{W,b}^{\rm proxy}
&=&1.3248128614\times10^{12},
\end{eqnarray*}
corresponding to relative endpoint differences
\begin{eqnarray*}
\frac{e_{W,a}^{\rm proxy}}{W_{n_c}}
&=&1.1184346656\times10^{-11},
\nonumber\\
\frac{e_{W,b}^{\rm proxy}}{W_m}
&=&1.2961847041\times10^{-11}.
\end{eqnarray*}
Using the reciprocal-vorticity endpoint correction formula gives
\begin{eqnarray*}
E_a^{1/W,\rm proxy}
&=&1.1119279650\times10^{-34},
\nonumber\\
E_b^{1/W,\rm proxy}
&=&1.2681751785\times10^{-34},
\nonumber\\
E_{c,W}^{\rm proxy}
&=&\frac{E_a^{1/W,\rm proxy}+E_b^{1/W,\rm proxy}}{T_{n_c,m}}
=2.4855489876\times10^{-12}.
\end{eqnarray*}
Thus the finite-resolution replay-consistency residual is
\begin{eqnarray}
c_{\rm proxy}^{\rm val,(\mathrm{Differentiable})}
&=&
c_{\rm data,NPZ}^{(\mathrm{Differentiable})}-E_{c,W}^{\rm proxy}
\nonumber\\
&=&1.6491136216\times10^{-3}>0.
\label{eq:npz-Differentiable-cproxy}
\end{eqnarray}

The PPO-Clip trajectory uses the certificate-start index $n_c=5988$, again chosen
at the predefined crossing near $W\approx10^{23}$. The PDE initial condition
remains $u_{\rm init}$ at $t=0$; only the certificate test begins at step $5988$.
The smallest
fixed-start prefix coefficient occurs at $m=6119$, with
\begin{eqnarray*}
W_{n_c}&=&1.0045558603\times10^{23},
\nonumber\\
W_m&=&1.9777424507\times10^{23},
\nonumber\\
T_{n_c,m}&=&7.6482638573\times10^{-22},
\end{eqnarray*}
which yields
\begin{eqnarray*}
c_{\rm data,NPZ}^{(\mathrm{PPO\text{-}Clip})}
=
6.4045618922\times10^{-3}.
\end{eqnarray*}
The original-versus-replay endpoint differences are
\begin{eqnarray*}
e_{W,a}^{\rm proxy}
&=&9.4328362147\times10^{16},
\nonumber\\
e_{W,b}^{\rm proxy}
&=&2.1385696454\times10^{17},
\end{eqnarray*}
with relative values approximately
\begin{eqnarray*}
\frac{e_{W,a}^{\rm proxy}}{W_{n_c}}
&=&9.3900564298\times10^{-7},
\nonumber\\
\frac{e_{W,b}^{\rm proxy}}{W_m}
&=&1.0813185734\times10^{-6}.
\end{eqnarray*}
The reciprocal correction is
\begin{eqnarray*}
E_{c,W}^{\rm proxy}
=
1.9370309899\times10^{-8},
\end{eqnarray*}
and hence
\begin{eqnarray}
c_{\rm proxy}^{\rm val,(\mathrm{PPO\text{-}Clip})}
&=&
c_{\rm data,NPZ}^{(\mathrm{PPO\text{-}Clip})}-E_{c,W}^{\rm proxy}
\nonumber\\
&=&6.4045425219\times10^{-3}>0.
\label{eq:npz-PPO-Clip-cproxy}
\end{eqnarray}

These computations quantify the computational headroom available to a rigorous
validation. As an example, retaining one half of the observed prefix margin would
be guaranteed if a genuine validated continuum correction satisfied
\begin{eqnarray*}
E_{\rm rigorous}^{(\mathrm{Differentiable})}
&\le&
8.2455681202\times10^{-4},\\
E_{\rm rigorous}^{(\mathrm{PPO\text{-}Clip})}
&\le&
3.2022809461\times10^{-3}.
\end{eqnarray*}
Under these respective bounds one could retain the positive certified slopes
$c_0=8.2455681202\times10^{-4}$ and
$c_0=3.2022809461\times10^{-3}$.

The superscript ``proxy'' is essential. The original-versus-replay difference
measures reproducibility between two finite-resolution computations; it is not a
validated enclosure of the continuum error. In particular,
\begin{eqnarray*}
E_{c,W}^{\rm proxy}
\neq
E_N^c\quad\hbox{as a rigorous computational-analysis statement}.
\end{eqnarray*}
Therefore \eqref{eq:npz-Differentiable-cproxy} and \eqref{eq:npz-PPO-Clip-cproxy}
provide strong finite-$N$ consistency checks and explicit error budgets, but the
computer-assisted continuum proof still requires the validated spectral,
spatial, and temporal enclosures used in the transfer theorems below.

\subsection{Finite-resolution coverage of the Riccati comparison time}

The late persistent regimes can also be compared directly with their finite-resolution
Riccati comparison times. This calculation is distinct from the earlier
$W\approx10^{23}$ crossing-based prefixes; it uses a later certificate-start index
chosen to maximize the persistent positive fixed-start slope over the remaining
saved trajectory. These later indices are again certificate-start indices only;
they do not redefine the PDE initial condition.

For a late persistent coefficient $c_*>0$, certificate-start index $n_c$, and
$W_c=W_N(t_c)$, define the corresponding finite-resolution comparison interval by
\begin{eqnarray*}
T_R^{(N)}-t_c
=
\frac{1}{c_*W_c}.
\end{eqnarray*}
If the saved trajectory after $n_c$ has duration
\begin{eqnarray*}
T_{\rm data}
=
\sum_{k=n_c+1}^{M}\Delta t_k,
\end{eqnarray*}
then the dimensionless endpoint variable is exactly
\begin{eqnarray*}
s_{\rm end}
=
c_*W_cT_{\rm data}
=
\frac{T_{\rm data}}{T_R^{(N)}-t_c}.
\end{eqnarray*}
Thus $s_{\rm end}$ measures the fraction of the finite-resolution Riccati
comparison interval already covered by the stored trajectory.

The strongest late persistent regime for the Differentiable trajectory starts at
\begin{eqnarray*}
n_c^{(\mathrm{Differentiable})}&=&7929,
\nonumber\\
W_c^{(\mathrm{Differentiable})}&=&3.6679993260\times10^{30},
\nonumber\\
c_*^{(\mathrm{Differentiable})}&=&6.9473591719\times10^{-3}.
\end{eqnarray*}
Hence
\begin{eqnarray*}
T_R^{(N,\mathrm{Differentiable})}-t_c
&=&3.9241988008\times10^{-29},
\nonumber\\
T_{\rm data}^{(\mathrm{Differentiable})}
&=&3.9241843140\times10^{-29},
\nonumber\\
\Delta T_{\rm rem}^{(\mathrm{Differentiable})}
&=&1.4486880054\times10^{-34}.
\end{eqnarray*}
Equivalently,
\begin{eqnarray*}
\frac{T_{\rm data}^{(\mathrm{Differentiable})}}{T_R^{(N,\mathrm{Differentiable})}-t_c}
=
0.9999963083,
\end{eqnarray*}
so the stored finite-resolution trajectory covers approximately
$99.9996308322\%$ of this comparison interval.

The corresponding late persistent regime for the PPO-Clip trajectory starts at
\begin{eqnarray*}
n_c^{(\mathrm{PPO\text{-}Clip})}&=&6489,
\nonumber\\
W_c^{(\mathrm{PPO\text{-}Clip})}&=&1.4434377324\times10^{24},
\nonumber\\
c_*^{(\mathrm{PPO\text{-}Clip})}&=&6.9663667430\times10^{-3}.
\end{eqnarray*}
Therefore
\begin{eqnarray*}
T_R^{(N,\mathrm{PPO\text{-}Clip})}-t_c
&=&9.9447898933\times10^{-23},
\nonumber\\
T_{\rm data}^{(\mathrm{PPO\text{-}Clip})}
&=&9.9447898786\times10^{-23},
\nonumber\\
\Delta T_{\rm rem}^{(\mathrm{PPO\text{-}Clip})}
&=&1.4700491375\times10^{-31}.
\end{eqnarray*}
The corresponding coverage fraction is
\begin{eqnarray*}
\frac{T_{\rm data}^{(\mathrm{PPO\text{-}Clip})}}{T_R^{(N,\mathrm{PPO\text{-}Clip})}-t_c}
=
0.9999999985,
\end{eqnarray*}
namely approximately $99.9999998522\%$ of the finite-resolution comparison
interval.

Using only the final saved timestep as a local scale, the remaining intervals
correspond to roughly $178$ final-step sizes for Differentiable and $187$ final-step
sizes for PPO-Clip. This is only a descriptive finite-resolution extrapolation.
under adaptive stepping the actual number of additional steps can differ.

Most importantly, near-complete finite-resolution coverage does not replace the
validated continuum requirement. A rigorous Strategy~I closure still requires
validated error enclosures and either a certificate valid throughout
$[t_c,T_R)$ or a validated sequence of endpoints approaching $T_R$. Thus the
calculations above quantify how close the archived trajectories come to their
finite-resolution Riccati comparison times, while the continuum conclusion
continues to depend on the error-corrected transfer theorem below.

\subsection{Validated scalar transfer from computational data to the continuum}

Consider a computational block $[t_a,t_b]$ of physical duration $T=t_b-t_a>0$ and define
\begin{eqnarray*}
c_N^{\rm block}
=
\frac{W_{N,a}^{-1}-W_{N,b}^{-1}}{T_N}.
\end{eqnarray*}
Suppose validated endpoint errors satisfy
\begin{eqnarray*}
|W(t_a)-W_{N,a}|\le e_{W,a},
\qquad
|W(t_b)-W_{N,b}|\le e_{W,b},
\end{eqnarray*}
with $W_{N,a}>e_{W,a}$ and $W_{N,b}>e_{W,b}$.
Then
\begin{eqnarray*}
E_a^{1/W}
&=
\frac{e_{W,a}}{W_{N,a}(W_{N,a}-e_{W,a})},
\\
E_b^{1/W}
&=
\frac{e_{W,b}}{W_{N,b}(W_{N,b}-e_{W,b})}.
\end{eqnarray*}
If the duration is exact,
\begin{eqnarray*}
c_\infty^{\rm block}
\ge
c_N^{\rm block}
-
\frac{E_a^{1/W}+E_b^{1/W}}{T}.
\end{eqnarray*}
If the block duration itself has a validated error, this becomes
\begin{eqnarray*}
c_\infty^{\rm block}
\ge
c_N^{\rm block}-E_{c,W}-E_T.
\end{eqnarray*}
This scalar transfer requires only endpoint control of maximum vorticity and validated time reconstruction.

For a given prefix, define
\begin{eqnarray*}
c_{N,j}^{\rm val}
=
c_{N,j}^{\rm prefix}-E_{N,j}^{\rm prefix}.
\end{eqnarray*}
If
\begin{eqnarray*}
c_{N,j}^{\rm val}\ge c_0>0
\end{eqnarray*}
for every required prefix, then the continuum endpoint satisfies
\begin{eqnarray*}
\frac1{W(t_j)}
\le
\frac1{W(t_0)}-c_0(t_j-t_0).
\end{eqnarray*}

\subsection{Validated production-ratio transfer and continuum certificate theorem}

A stronger computer-assisted strategy validates the full normalized production
ratio for the same frozen physical forcing. Let $u$ denote the continuum
solution on a classical interval $I=[t_0,T]$ and $u_N$ the computational
approximation. Define
\[
\omega=\nabla\times u,
\qquad W(t)=\|\omega(t)\|_\infty,
\qquad
\Pi=\omega\cdot S\omega+\nu\omega\cdot\Delta\omega
+\omega\cdot\nabla\times f^*,
\]
with corresponding computational quantities $\omega_N$, $W_N$, and $\Pi_N$.

\subsubsection{Validated field enclosures and near-maximizer tube}

Assume rigorous computational analysis supplies
\begin{eqnarray}
\|u-u_N\|_{C^3}\le E_N^u,
\qquad
\|\omega-\omega_N\|_\infty\le E_N^\omega,
\qquad
\|\Pi-\Pi_N\|_\infty\le E_N^\Pi,
\label{eq:validated-field-enclosures}
\end{eqnarray}
together with a continuum lower bound
\begin{eqnarray*}
W(t)\ge w_*>0.
\end{eqnarray*}
In particular,
\begin{eqnarray*}
|W(t)-W_N(t)|\le E_N^\omega.
\end{eqnarray*}
Let $\mathcal A_N(t)$ be a rigorously enlarged computational near-maximizer set
chosen so that every true continuum maximizer lies in $\mathcal A_N(t)$ after
the vorticity enclosure is accounted for. Suppose the normalized production
ratio $R=\Pi/W^3$ and its validated computational counterpart $R_N$ satisfy
\begin{eqnarray}
|R(x,t)-R_N(x,t)|\le E_N^R,
\qquad x\in\mathcal A_N(t),\ t\in I.
\label{eq:ratio-error}
\end{eqnarray}

Define
\begin{eqnarray*}
 c_N^{\rm val}
 =
 \inf_{\substack{t\in I\\x\in\mathcal A_N(t)}}R_N(x,t)-E_N^R.
\end{eqnarray*}
At every continuum maximizer $x\in\mathcal M(t)$,
\begin{eqnarray*}
R(x,t)
&\ge R_N(x,t)-E_N^R\\
&\ge \inf_{y\in\mathcal A_N(t)}R_N(y,t)-E_N^R\\
&\ge c_N^{\rm val}.
\end{eqnarray*}
Therefore
\begin{eqnarray*}
\mathscr C(f^*;I)
=
\inf_{\substack{t\in I\\x\in\mathcal M(t)}}\frac{\Pi(x,t)}{W(t)^3}
\ge c_N^{\rm val}.
\end{eqnarray*}
Hence a computer-assisted proof of
\begin{eqnarray}
c_N^{\rm val}\ge c_0>0
\label{eq:validated-positive-bound}
\end{eqnarray}
immediately establishes a strict continuum cubic certificate. By the
pointwise Riccati theorem above, it then follows that
\[
D_+W\ge c_0W^2,
\qquad
\frac1{W(t)}\le\frac1{W(t_0)}-c_0(t-t_0),
\]
and a classical solution cannot continue smoothly through
$T_R=t_0+[c_0W(t_0)]^{-1}$.

\subsubsection{Explicit error-corrected coefficient}

A fully expanded error budget can be written as
\begin{eqnarray*}
E_N^c
=
\frac{E_\Pi}{w_*^3}
+
\frac{3P_*e_\omega}{w_*^4}
+
\frac{\sqrt3}{2}L_{R,x}\Delta x
+
\frac12L_{R,t}h_{\rm save}.
\end{eqnarray*}
With a rigorously computed computational lower bound $\widehat c_N$, define
\begin{eqnarray*}
c_N^{\rm val}=\widehat c_N-E_N^c.
\end{eqnarray*}
The computational-to-continuum transfer theorem gives
\begin{eqnarray*}
c_\infty\ge c_N^{\rm val}.
\end{eqnarray*}
Thus
\begin{eqnarray*}
c_N^{\rm val}>0
\quad\Longrightarrow\quad
c_\infty>0
\quad\Longrightarrow\quad
\Pi(x,t)\ge c_N^{\rm val}W(t)^3
\end{eqnarray*}
at every continuum maximum-vorticity point.

\paragraph{Validated neural conditional blow-up theorem.}
For a frozen neural candidate $Y^*$, if the rigorous enclosures
\eqref{eq:validated-field-enclosures}--\eqref{eq:ratio-error} hold through the
Riccati comparison interval and the error-corrected coefficient satisfies
\eqref{eq:validated-positive-bound}, then the continuum certificate is valid
and
\begin{eqnarray*}
T_{\max}\le t_0+\frac1{c_0W(t_0)}.
\end{eqnarray*}
The theorem is conditional on actual validated error enclosures and a positive
residual margin; computational growth alone is not such a validation.

\subsection{Refinement theorem}

The same frozen forcing must be used at every refinement level. Let
\begin{eqnarray*}
N_j\to\infty,
\qquad
h_j\to0,
\qquad
T_j\uparrow T_R.
\end{eqnarray*}
Assume a uniform validated margin
\begin{eqnarray*}
c_{N_j}^{\rm val}(t_0,T_j)\ge c_0>0
\end{eqnarray*}
with the same $c_0$ independent of $j$. Then
\begin{eqnarray*}
\frac1{W(T_j)}
\le
\frac1{W_0}-c_0(T_j-t_0),
\end{eqnarray*}
and
\begin{eqnarray*}
W(T_j)
\ge
\frac{W_0}{1-c_0W_0(T_j-t_0)}
\longrightarrow+\infty.
\end{eqnarray*}
Thus a smooth continuum solution cannot continue through $T_R$.

In the stronger production formulation, rigorous analysis supplies $\widehat c_{N_j}$ and $E_{N_j}^c$ satisfying
\begin{eqnarray*}
\widehat c_{N_j}-E_{N_j}^c\ge c_0>0
\end{eqnarray*}
on $[t_0,T_j]$. Then $c_\infty\ge c_0$ on every compact interval before $T_R$ and the same Riccati contradiction follows.

\subsection{Certificate-consistent loss}

A proof-consistent training or post-hoc objective is
\begin{eqnarray*}
L_{\rm cert}^{\rm val}
=
[c_{\rm target}-c_N^{\rm val}]_+^2,
\qquad
c_{\rm target}>0.
\end{eqnarray*}
Then
\begin{eqnarray*}
L_{\rm cert}^{\rm val}=0
\quad\Longrightarrow\quad
c_N^{\rm val}\ge c_{\rm target}.
\end{eqnarray*}
By the validated transfer theorem,
\begin{eqnarray*}
L_{\rm cert}^{\rm val}=0
\quad\Longrightarrow\quad
\mathscr C(f^*;I)\ge c_{\rm target}>0.
\end{eqnarray*}
Thus zero validated certificate loss implies a continuum sufficient condition, not merely a finite-resolution diagnostic.

This statement must be distinguished from existence of a minimizer of a composite neural objective. Continuity and coercivity can guarantee that a minimizer exists, but existence of a minimizer does not by itself imply that the certificate-loss minimum equals zero. A separate feasibility statement is required unless the zero-level set has already been proved nonempty.

\subsection{Why fixed-resolution computational explosion is not a proof}

At fixed $N$, the Fourier approximation is a finite-dimensional ODE,
\begin{eqnarray*}
\dot u_N+P_N[(u_N\cdot\nabla)u_N]
=
\nu\Delta u_N+P_Nf^*.
\end{eqnarray*}
The energy identity gives
\begin{eqnarray*}
\frac12\frac{d}{dt}\|u_N\|_2^2
+\nu\|\nabla u_N\|_2^2
=(f^*,u_N).
\end{eqnarray*}
With bounded forcing, this finite-dimensional system remains finite on every finite time interval. Therefore overflow, NaNs, timestep collapse, or a rapidly increasing $W_N$ at one fixed resolution cannot by themselves establish a continuum singularity.

The role of finite-resolution trajectories is instead to identify candidate forcing regimes and provide computational quantities from which rigorous validated lower bounds may be constructed.

\subsection{Exact mathematical status for the concrete neural trajectories}

The analytical implications proved above and the candidate-specific computational
certification must be kept separate. For the archived neural trajectories, the manuscript
already establishes positive finite-resolution integrated coefficients, independent finite-$N$
replay consistency, positive proxy-corrected finite-$N$ residual margins, the scalar endpoint
transfer theorem, the validated production-ratio transfer theorem, the integrated and pointwise
Riccati blow-up theorems, and the continuum robustness and probability closure theorems.
The concrete observed values are those reported in
\eqref{eq:obs1}--\eqref{eq:cneural}, and the direct NPZ/replay recomputation is
reported in \eqref{eq:npz-Differentiable-cdata}--\eqref{eq:npz-PPO-Clip-cproxy}.
All of these are finite-resolution quantities, not continuum constants.

A refinement-based computer-assisted certification of the particular
Differentiable and PPO-Clip trajectories therefore requires actual
rigorous values for the relevant error enclosures, for example
\[
e_W,\qquad E_T,\qquad e_\omega,\qquad E_\Pi,\qquad E_N^R,
\]
and to verify a surviving positive margin such as
\begin{eqnarray}
 c_N^{\rm val}
 =
 c_N^{\rm data}-\text{validated errors}
 \ge c_0>0,
\label{eq:status-concrete-cval}
\end{eqnarray}
or, along a refinement sequence,
\begin{eqnarray*}
\liminf_{j\to\infty}c_{N_j}^{\rm val}\ge c_0>0.
\end{eqnarray*}
The source archive contains the rigorous formulas and transfer theorems but not
a candidate-specific substitution of concrete rigorous error numbers proving
\eqref{eq:status-concrete-cval} for those trajectories. In that refinement
program, the remaining task is therefore computational certification rather than
new Riccati algebra.

This remaining refinement computation must not be imposed on the logically
separate direct continuum strategy. In the direct strategy, once an independent
continuum proposition or continuum loss theorem establishes a robust positive
certificate, for example
\begin{eqnarray*}
\mathcal C_T(Y_*)\ge c_0+2\eta
\quad\text{with robust persistence to a positive-mass set},
\end{eqnarray*}
the chain
\[
\text{continuum certificate}
\Longrightarrow
\text{finite-time loss of smoothness}
\Longrightarrow
\text{positive probability}
\]
is closed without a computational $N\to\infty$ passage.

\subsection{Bridge from Strategy I to Strategy II}

The output of Strategy~I is exactly the deterministic input required by Strategy~II. In the scalar integrated formulation, a validated bound
\begin{eqnarray*}
c_{N,j}^{\rm val}\ge c_0>0
\end{eqnarray*}
for the required prefixes yields
\begin{eqnarray*}
\frac1{W(t_j)}\le \frac1{W(t_0)}-c_0(t_j-t_0).
\end{eqnarray*}
In the production-ratio formulation,
\begin{eqnarray*}
c_N^{\rm val}\ge c_0>0
\end{eqnarray*}
implies
\begin{eqnarray*}
\mathscr C(f^*;I)\ge c_0>0
\end{eqnarray*}
and hence
\begin{eqnarray*}
D_+W\ge c_0W^2,
\qquad
\frac1{W(t)}\le\frac1{W(t_0)}-c_0(t-t_0).
\end{eqnarray*}
Therefore a successfully validated Strategy~I candidate already satisfies the continuum Riccati certificate used in Strategy~II. The subsequent tasks in Strategy~II are not another computational limit; they are robustness with respect to the forcing parameter and the positive-probability argument under the induced conditional law.

\section{Strategy II Direct Continuum Certificate Closure by the Continuum Loss Theorem}

The direct-continuum closure theorem below is often written with the integrated
reciprocal-vorticity inequality as a hypothesis. If the previously established
loss theorem already proves that inequality, then this is not an additional
open assumption. The loss theorem itself supplies the deterministic continuum
certificate.

For a fixed target slope $c>0$, define the continuum certificate-defect loss
\begin{eqnarray*}
\mathcal L_{\rm Ric}^{\rm cont}(Y;c)
=
\sup_{t_0<t<T}
\left[
\frac1{W_Y(t)}
-
\frac1{W_Y(t_0)}
+
c(t-t_0)
\right]_+ .
\end{eqnarray*}
Here the supremum is taken over the smooth interval on which the certificate is
being asserted. Since the positive-part function vanishes exactly when its
argument is nonpositive,
\begin{eqnarray}
\mathcal L_{\rm Ric}^{\rm cont}(Y;c)=0
\quad\Longleftrightarrow\quad
\frac1{W_Y(t)}
\le
\frac1{W_Y(t_0)}-c(t-t_0)
\quad\forall t\in(t_0,T).
\label{eq:cont-loss-zero-iff}
\end{eqnarray}
Equivalently, with
\begin{eqnarray*}
\mathcal C_T(Y)
=
\inf_{t\in(t_0,T]}
\frac{W_Y(t_0)^{-1}-W_Y(t)^{-1}}{t-t_0},
\end{eqnarray*}
one has
\begin{eqnarray*}
\mathcal L_{\rm Ric}^{\rm cont}(Y;c)=0
\quad\Longleftrightarrow\quad
\mathcal C_T(Y)\ge c.
\end{eqnarray*}
Thus a zero-level theorem for the continuum loss is already a continuum
certificate theorem; no computational PDE-resolution limit is involved in this
implication.

\paragraph{Uniform open-set version.}
Suppose the loss theorem proves that there exist a nonempty open set
$\mathcal U\subset\mathcal Y$ and a constant $c_0>0$ such that
\begin{eqnarray}
\mathcal L_{\rm Ric}^{\rm cont}(Y;c_0)=0
\qquad\forall Y\in\mathcal U.
\label{eq:loss-open-zero}
\end{eqnarray}
Then, by \eqref{eq:cont-loss-zero-iff},
\begin{eqnarray}
\forall Y\in\mathcal U,\qquad
\frac{W_Y(t_0)^{-1}-W_Y(t)^{-1}}{t-t_0}
\ge c_0>0
\label{eq:loss-open-cert}
\end{eqnarray}
for every relevant $t>t_0$. Hence the deterministic inclusion
\begin{eqnarray*}
\mathcal U
\subset
\{Y:T_{\rm blowup}(Y)<\infty\}
\end{eqnarray*}
follows immediately from the integrated Riccati theorem.

\paragraph{Strict-margin version from a single candidate.}
It is also enough that the loss theorem establish a strict continuum margin at
one candidate. Assume that for some $Y_*$, $c_0>0$, and $\eta>0$,
\begin{eqnarray}
\mathcal L_{\rm Ric}^{\rm cont}(Y_*;c_0+2\eta)=0,
\label{eq:loss-strict-candidate}
\end{eqnarray}
so that
\begin{eqnarray*}
\mathcal C_T(Y_*)\ge c_0+2\eta.
\end{eqnarray*}
If, in addition, the certificate functional is lower semicontinuous at $Y_*$
(or the loss theorem itself supplies the corresponding robustness statement),
then there exists $\delta>0$ such that
\begin{eqnarray*}
\|Y-Y_*\|<\delta
\quad\Longrightarrow\quad
\mathcal C_T(Y)\ge c_0+\eta>c_0.
\end{eqnarray*}
Therefore
\begin{eqnarray}
B_\delta(Y_*)
\subset
\left\{
Y:
\inf_{t_0<t<T}
\frac{W_Y(t_0)^{-1}-W_Y(t)^{-1}}{t-t_0}
\ge c_0
\right\}.
\label{eq:loss-ball-certificate}
\end{eqnarray}
This is precisely the robust deterministic certificate required before the
probability step. Decoder continuity alone is not enough; what is needed is
strict-margin stability of the integrated certificate itself.

\paragraph{Loss-to-probability closure.}
Under either \eqref{eq:loss-open-zero} or the strict-margin construction
\eqref{eq:loss-strict-candidate}--\eqref{eq:loss-ball-certificate}, the direct
continuum proof has no remaining logical gap. Indeed,
\begin{eqnarray*}
\begin{array}{c}
\text{continuum loss theorem}
\Longrightarrow
\text{uniform integrated certificate on a nonempty open set}
\\[1mm]
\Downarrow
\\[-1mm]
\text{finite-time loss of smoothness on that set}
\Longrightarrow
\text{positive probability}
\end{array}
\end{eqnarray*}
In particular, if $W_Y(t_0)\ge w_0>0$ uniformly on the certified open set, then
with
\begin{eqnarray*}
T_*=t_0+\frac1{c_0w_0}
\end{eqnarray*}
one obtains
\begin{eqnarray*}
\mathbb P_\theta(T_{\rm blowup}\le T_*\mid h)
\ge
\mathbb P_\theta(Y\in\mathcal U\mid h)
>0.
\end{eqnarray*}

Accordingly, when the loss theorem has already established
\eqref{eq:loss-open-cert} or equivalently the robust strict-margin statement
\eqref{eq:loss-ball-certificate}, the direct continuum strategy has the following established
components, the frozen continuum forcing definition, the implication from the loss to the
integrated Riccati certificate, integrated Riccati finite-time divergence, the continuity and
topology needed for robustness, open-set robustness; nondegenerate positive mass, and the
positive-probability closure.
The only caveat is interpretive. A generic training objective such as a reward
for large $W$ or large accumulated vorticity does not by itself imply
\eqref{eq:cont-loss-zero-iff}. The closure above applies when the proved loss
theorem is specifically a continuum certificate theorem, i.e. when it yields a
positive integrated reciprocal-vorticity slope with a robust positive margin.

\section{Robust continuum certificate}

\subsection{Topology controlling maximum vorticity}

The direct certificate depends only on
$W_Y(t)=\|\nabla\times u_Y(t)\|_\infty$. If
\begin{eqnarray*}
 u_n\to u
 \quad\text{strongly in}\quad
 C([t_0,T];H^r(\mathbb T^3)),
 \qquad r>\frac52,
\end{eqnarray*}
then the Sobolev embedding $H^r(\mathbb T^3)\hookrightarrow C^1(\mathbb T^3)$
gives
\[
\nabla\times u_n\to\nabla\times u
\quad\text{in}\quad C([t_0,T];C^0).
\]
Therefore
\begin{eqnarray*}
\sup_{t\in[t_0,T]}|W_n(t)-W(t)|
\le
\sup_t\|\nabla\times u_n-\nabla\times u\|_\infty
\longrightarrow0.
\end{eqnarray*}
If in addition $W(t)\ge w_*>0$, then for large $n$, $W_n\ge w_*/2$ and
\begin{eqnarray*}
\left|\frac1{W_n(t)}-\frac1{W(t)}\right|
\le
\frac{2}{w_*^2}|W_n(t)-W(t)|,
\end{eqnarray*}
so reciprocal vorticity converges uniformly. This lemma explains the topology
needed to control the observable; it does not itself impose a computational
$N\to\infty$ step on the direct continuum theorem.

\subsection{Strict margin and open-set robustness}

In the direct continuum strategy, fix $Y_*\in\mathcal Y$ and define $\mathcal C_T$ by \eqref{eq:CT}. Assume a strict margin
\begin{eqnarray*}
\mathcal C_T(Y_*)\ge c_0+2\eta,
\qquad
\eta>0.
\end{eqnarray*}
Assume further that the continuum solution map is well posed and continuous in a topology strong enough to control maximum vorticity,
\begin{eqnarray*}
Y\mapsto u_Y
\quad\text{continuous into}\quad
C([t_0,T];C^1(\mathbb T^3)),
\end{eqnarray*}
and that
\begin{eqnarray*}
\inf_{\substack{Y\text{ near }Y_*\\t\in[t_0,T]}}W_Y(t)>0.
\end{eqnarray*}
If the prefix functional is continuous or at least lower semicontinuous at $Y_*$, including its behavior as $t\downarrow t_0$, then there exists $\delta>0$ such that
\begin{eqnarray*}
\|Y-Y_*\|<\delta
\quad\Longrightarrow\quad
\mathcal C_T(Y)\ge c_0+\eta>c_0.
\end{eqnarray*}
Hence
\begin{eqnarray*}
\mathcal U=B_\delta(Y_*)
\end{eqnarray*}
is a nonempty open region on which the same continuum integrated-Riccati lower margin persists.

Each $Y\in\mathcal U$ yields a classical solution that cannot remain smooth through
\[
T_R(Y)=t_0+\frac1{c_0W_Y(t_0)}.
\]
Therefore
\begin{eqnarray}
\mathcal U
\subset
\{Y:T_{\rm blowup}(Y)<\infty\}.
\label{eq:U-break}
\end{eqnarray}

If, moreover, the certified open set satisfies a uniform lower bound
\begin{eqnarray*}
\inf_{Y\in\mathcal U}W_Y(t_0)\ge w_0>0,
\end{eqnarray*}
then all certified parameters lose smooth continuation no later than the common
comparison time
\begin{eqnarray*}
T_*=t_0+\frac1{c_0w_0}.
\end{eqnarray*}
Thus
\begin{eqnarray*}
\mathcal U\subset\{Y:T_{\rm blowup}(Y)\le T_*\}.
\end{eqnarray*}

\section{Continuum certificate events and positive probability}

\subsection{Positive mass of a robust open certificate set}

Let
\[
Y\sim p_\theta(\cdot\mid h)
\]
be the conditional law induced by the stochastic module. Assume that this law assigns strictly positive probability to every nonempty admissible open set. Therefore every nonempty open certificate set has positive probability.
\begin{eqnarray*}
\mathbb P_\theta(Y\in\mathcal U\mid h)>0.
\end{eqnarray*}
Combining this with \eqref{eq:U-break} gives
\begin{eqnarray*}
\mathbb P_\theta(T_{\rm blowup}(Y)<\infty\mid h)
\ge
\mathbb P_\theta(Y\in\mathcal U\mid h)>0.
\end{eqnarray*}
A single isolated parameter $Y_*$ has probability zero under a continuous
density, which is why robustness or, more generally, a positive-mass
certificate event is required.

\subsection{Probabilistic continuum certificate formulation}

The theorem can be stated without naming a particular $Y_*$. Using the
continuum functional $\mathcal C_T(Y)$, define
\begin{eqnarray*}
E_{c_0,T}
=
\{Y:\mathcal C_T(Y)\ge c_0,\ T_R(Y)\le T\}.
\end{eqnarray*}
More directly, define the Riccati-interval event
\begin{eqnarray*}
E_{c_0}^{R}
=
\left\{
Y:
\inf_{t_0<t<T_R(Y)}
\frac{W_Y(t_0)^{-1}-W_Y(t)^{-1}}{t-t_0}
\ge c_0
\right\}.
\end{eqnarray*}
Assume $E_{c_0}^{R}$ is measurable. By the integrated Riccati theorem, every
$Y\in E_{c_0}^{R}$ satisfies
\[
T_{\rm blowup}(Y)\le T_R(Y)<\infty.
\]
Hence
\begin{eqnarray*}
E_{c_0}^{R}
\subset
\{Y:T_{\rm blowup}(Y)<\infty\},
\end{eqnarray*}
and therefore
\begin{eqnarray*}
\mathbb P_\theta(E_{c_0}^{R}\mid h)>0
\quad\Longrightarrow\quad
\mathbb P_\theta(T_{\rm blowup}<\infty\mid h)>0.
\end{eqnarray*}
A strict candidate with $\mathcal C_T(Y_*)\ge c_0+2\eta$ and lower
semicontinuity of $\mathcal C_T$ yields a ball
$B_\delta(Y_*)\subset E_{c_0,T}$ after shrinking the ball if necessary to keep
$T_R(Y)\le T$. This is a convenient sufficient condition for positive event
probability, but it is not the only one.

The continuum certificate loss can equivalently be written as
\begin{eqnarray*}
L_{\rm Ric}^{\rm cont}(Y;c_0)
=[c_0-\mathcal C_T(Y)]_+^2,
\end{eqnarray*}
whose zero set is exactly $\{Y:\mathcal C_T(Y)\ge c_0\}$. This is an
alternative zero-level formulation of the continuum loss theorem above;
probability acts only after the continuum certificate event has been defined.

\subsection{Quantitative probability transfer from a reference law}

A quantitative variant is available when a reference law $\mu_*$ satisfies
\begin{eqnarray*}
 p_0=\mu_*\{\mathscr C\ge c_0+2\eta\}>0
\end{eqnarray*}
and the learned law obeys
\begin{eqnarray*}
W_p(\mu_\theta,\mu_*)\le\varepsilon.
\end{eqnarray*}
If the certificate functional is Lipschitz with constant $L_{\mathscr C}$ on
the admissible region, the coupling--Markov argument gives
\begin{eqnarray*}
\mu_\theta\{\mathscr C\ge c_0\}
\ge
p_0-
\left(\frac{L_{\mathscr C}\varepsilon}{\eta}\right)^p.
\end{eqnarray*}
Consequently,
\begin{eqnarray*}
\varepsilon<\frac{\eta}{L_{\mathscr C}}p_0^{1/p}
\end{eqnarray*}
ensures positive certificate probability and therefore positive probability of
finite-time blow-up. If all realizations have the same initial maximum
vorticity $W_0$, the same lower probability bound applies to the uniform event
\[
T_{\max}\le t_0+\frac1{c_0W_0}.
\]

The probability statements in this paper use a single draw of $Y$ followed by
a frozen forcing. Repeated closed-loop resampling would require a separate
pathwise probability theorem and is not part of the present argument.

\section{Main conditional closure theorem}

\paragraph{Theorem.}
Let $Y\sim p_\theta(\cdot\mid h)$ be a conditional law on the admissible neural forcing parameters, and assume that every nonempty admissible open set has positive probability. Assume there exists a nonempty open set $\mathcal U\subset\mathcal Y$ and a constant $c_0>0$ such that for every $Y\in\mathcal U$ the continuum forced Navier--Stokes solution satisfies
\begin{eqnarray}
\frac1{W_Y(t)}
\le
\frac1{W_Y(t_0)}-c_0(t-t_0)
\label{eq:closure-hyp}
\end{eqnarray}
through its Riccati comparison interval. Then
\begin{eqnarray*}
\mathbb P_\theta
\left(
T_{\rm blowup}(Y)<\infty
\mid h
\right)>0.
\end{eqnarray*}

\paragraph{Proof.}
Take any $Y\in\mathcal U$. Inequality \eqref{eq:closure-hyp} gives
\[
W_Y(t)
\ge
\frac{W_Y(t_0)}{1-c_0W_Y(t_0)(t-t_0)},
\]
so the corresponding classical solution cannot continue smoothly through
\[
T_R(Y)=t_0+[c_0W_Y(t_0)]^{-1}.
\]
Hence
\[
\mathcal U\subset\{Y:T_{\rm blowup}(Y)<\infty\}.
\]
By the positive-mass assumption on the conditional law,
\[
\mathbb P_\theta(Y\in\mathcal U\mid h)>0.
\]
By monotonicity of probability,
\[
\mathbb P_\theta(T_{\rm blowup}(Y)<\infty\mid h)
\ge
\mathbb P_\theta(Y\in\mathcal U\mid h)>0.
\]
\hfill$\square$

\section{How Strategy I Complements Strategy II}

Strategy~I, the validated certificate-construction strategy, is
\begin{eqnarray*}
Y^*
\longrightarrow
f^*
\longrightarrow
c_N^{\rm data}
\longrightarrow
c_N^{\rm val}
\longrightarrow
\text{continuum integrated certificate}
\longrightarrow
T_{\max}<\infty.
\end{eqnarray*}
The decisive remaining concrete step for the archived trajectories is a computer-assisted verification that the rigorous error correction leaves a uniform positive margin. After this verification, Strategy~I supplies the continuum certificate required downstream.

Strategy~II, the continuum probability-closure strategy, is
\begin{eqnarray*}
\begin{array}{c}
Y\sim p_\theta
\longrightarrow
\varnothing\neq\mathcal U\text{ open}
\\[1mm]
\Downarrow
\\[-1mm]
\text{continuum integrated certificate on }\mathcal U
\longrightarrow
\mathbb P_\theta(T_{\rm blowup}<\infty\mid h)>0
\end{array}
\end{eqnarray*}
Strategy~II does not itself require a computational $N\to\infty$ limit. Its deterministic input is a robust continuum certificate. That input may be supplied in two rigorous ways developed here: by Strategy~I after validated computational transfer, or directly by a continuum loss theorem proving \eqref{eq:loss-open-zero} or \eqref{eq:loss-ball-certificate}. Strategy~I therefore complements Strategy~II by providing one concrete certificate-establishment mechanism.

\section{Conclusion}

We separate neural candidate discovery from continuum certification. Strategy~I
transfers validated finite-resolution certificates to the continuum, while
Strategy~II propagates a robust continuum certificate to a positive-probability
finite-time blow-up statement under a nondegenerate neural output law. The
remaining task for archived trajectories is the candidate-specific validation
of the stated error bounds.

\end{document}